\documentclass[10pt,twocolumn,letterpaper]{article} 
   \usepackage{times,epsfig,amssymb,amstex,subfigure}

\date{}

\title{{\bf Differential Invariants under Gamma Correction}}
\author{Andreas Siebert \\
	Department of Computer Science \\
	The University of British Columbia \\
	siebert@@cs.ubc.ca}

\begin{document}

\maketitle

\begin{abstract}
{\it
This paper presents invariants under gamma correction
and similarity transformations.
The invariants are local features based on differentials
which are implemented using derivatives of the Gaussian.
The use of the proposed invariant representation is shown to yield
improved correlation results in a template matching scenario.
}
\end{abstract}

%****************************************************************************
\section{Introduction}
%****************************************************************************

Invariants are a popular concept in object recognition
and image retrieval~\cite{aw99, fmzchr91, mz92a, rw95, w93, wood96}.
They aim to provide descriptions
that remain constant under certain geometric or radiometric transformations
of the scene, thereby reducing the search space.
They can be classified into global invariants, typically based either on a set
of key points or on moments,
and local invariants,
typically based on derivatives of the image function
which is assumed to be continuous and differentiable.

The geometric transformations of interest often include
translation, rotation, and scaling,
summarily referred to as {\em similarity} transformations.
In a previous paper~\cite{s99}, building on work done by
Schmid and Mohr~\cite{sm97}, we have proposed differential invariants
for those similarity transformations, plus {\em linear} brightness change.
Here, we are looking at a {\em non-linear} brightness change
known as {\em gamma correction}.

Gamma correction is a non-linear quantization of the brightness measurements 
performed by many cameras
during the image formation process.\footnote{
Historically, the parameter gamma was introduced to describe the 
non-linearity of photographic film.
Today, its main use is to improve the output
of cathode ray tube based monitors,
but the gamma correction in {\em display} devices
is of no concern to us here.}
The idea is to achieve better {\em perceptual}
results by maintaining an approximately constant ratio
between adjacent brightness levels, placing the quantization levels
apart by the {\em just noticeable difference}.
Incidentally, this non-linear quantization also precompensates
for the non-linear mapping from voltage to brightness
in electronic display devices~\cite{h98b, poyn96}.

\begin{figure}[htbp]
 \begin{center}
  \epsfig{file=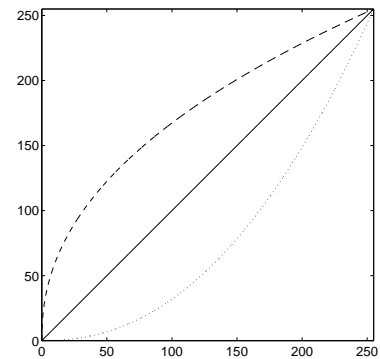,width=4.9cm}
  \caption
        {Gamma correction as a function of intensity.
        \ (solid)~ $\gamma=1$;
        \ (dashed)~$\gamma=0.45$;
        \ (dotted)~$\gamma=2.22$.
	Note how, for $\gamma<1$, the lower intensities are mapped
	onto a larger range.
        }
  \label{fig:GammaCorr}
 \end{center}
\end{figure}

Gamma correction can be expressed by the equation
\begin{equation} \label{eq:GammaCorr}
   I_\gamma = p\, I^\gamma
\end{equation}
where $I$ is the input intensity, $I_\gamma$ is the output intensity,
and $p$ is a normalization factor which is determined by the value of $\gamma$.
For output devices, the NTSC standard specifies $\gamma=2.22$.
For input devices like cameras, the parameter value is just inversed,
resulting in a typical value of $\gamma=1/2.22=0.45$.
The camera we used, the Sony 3 CCD color camera DXC~950,
exhibited $\gamma \approx 0.6$.\footnote{
Martin~\cite{mar93} reports the settings of $\gamma=0.45, 0.50, 0.60$
for the Kodak Megaplus XRC camera}
Fig.~\ref{fig:GammaCorr} shows the intensity mapping of 8-bit data for different
values of $\gamma$.

It turns out that an invariant under gamma correction 
can be designed from first and second order derivatives.
Additional invariance under scaling requires third order derivatives.
Derivatives are by nature translationally invariant.
Rotational invariance in 2-d is achieved by using rotationally symmetric
operators.

%****************************************************************************
\section{The Invariants}
%****************************************************************************

The key idea for the design of the proposed invariants is to form
suitable ratios of the derivatives of the image function such that
the parameters describing the transformation of interest will cancel out.
This idea has been used in~\cite{s99} to achieve invariance under linear
brightness changes, and it can be adjusted to the context of gamma correction
by -- at least conceptually -- considering the {\em logarithm} of
the image function.
For simplicity, we begin with 1-d image functions.

%--------------------------------------------------------------------------
\subsection{Invariance under Gamma Correction}
%--------------------------------------------------------------------------

Let $f(x)$ be the image function, i.e.~the original signal, assumed
to be continuous and differentiable, and
$f_\gamma(x)=p\,f(x)^\gamma$ the corresponding gamma corrected function.
Note that $f(x)$ is a special case of $f_\gamma(x)$ where $\gamma=p=1$.
Taking the logarithm yields
\begin{equation} \label{eq:fgammalog}
  \tilde{f}_\gamma(x) = \ln (p\,f(x)^\gamma) = \ln p + \gamma \ln f(x)
\end{equation}
with the derivatives
$\tilde{f}'_\gamma(x) = \gamma\,f'(x)/f(x)$, and
$\tilde{f}''_\gamma(x) = \gamma\,(f(x)\,f''(x) - f'(x)^2)/f(x)^2$.
We can now define the invariant $\Theta_{12\gamma}$ under gamma correction
to be
{\large
\begin{equation} \label{eq:Th12G}
 \begin{tabular}{ll}
   {\normalsize $\Theta_{12\gamma}(f(x))$} &=
                $\frac{\tilde{f}'_\gamma(x)} {\tilde{f}''_\gamma(x)}$ \\
        \rule[-6.0mm]{0mm}{13mm}
        &= $\frac{\gamma\, \frac{f'(x)}{f(x)}}
                {\gamma\, \frac{f(x)\,f''(x) - f'(x)^2}{f(x)^2}}$ \\
        &= $\frac{f(x)\,f'(x)}{f(x)\,f''(x) - f'(x)^2}$
 \end{tabular}
\end{equation}
}
The factor $p$ has been eliminated by taking derivatives, and
$\gamma$ has canceled out.
Furthermore, $\Theta_{12\gamma}$ turns out to be completely specified in terms
of the {\em original} image function and its derivatives,
i.e.~the logarithm actually doesn't have to be computed.
The notation $\Theta_{12\gamma}(f(x))$ indicates that the invariant
depends on the underlying image function $f(x)$ and location $x$ --
the invariance holds under gamma correction, not under spatial changes
of the image function.

A shortcoming of $\Theta_{12\gamma}$
is that it is undefined where the denominator is zero.
Therefore, we modify $\Theta_{12\gamma}$ to be continuous everywhere:
{\large
\begin{equation} \label{eq:Thm12G}
 \begin{tabular}{lcl}
       &0 & {\normalsize if $f\,f'=0 \wedge f\,f'' - {f'}^2=0$} \\
	 \rule[-4mm]{0mm}{8mm}
    {\normalsize $\Theta_{m12\gamma}=$}
       &$\frac{f\,f'}{f\,f'' - {f'}^2}$
          &{\normalsize if $|f\,f'| < |f\,f'' - {f'}^2|$} \\
       &$\frac{f\,f'' - {f'}^2}{f\,f'}$
          &{\normalsize else} \\
 \end{tabular}
\end{equation}
}where, for notational convenience, we have dropped the variable $x$.
The modification entails
\mbox{$-1 \leq \Theta_{m12\gamma} \leq 1$}.
Note that the modification is just a heuristic to deal with poles.
If all derivatives are zero because the image function is constant,
then differentials are certainly not the best way to represent the function.

%--------------------------------------------------------------------------
\subsection{Invariance under Gamma Correction and Scaling}
%--------------------------------------------------------------------------

If scaling is a transformation that has to be considered,
then another parameter $\alpha$ describing the change of size
has to be introduced.
That is, scaling is modeled here as variable substitution~\cite{sm97}:
the scaled version of $f(x)$ is $g(\alpha x)=g(u)$.
So we are looking at the function
\[
 \tilde{f}_\gamma(x) = \ln (p f(x)^\gamma) =
		      \ln p + \gamma \ln g(\alpha x) = \tilde{g}_\gamma(u)
\]
where the derivatives with respect to $x$ are
$\tilde{g}'_\gamma(u)= \gamma \alpha\, g'(u)/g(u)$,
$\tilde{g}''_\gamma(u)= \gamma \alpha^2 (g(u)\,g''(u) - g'(u)^2)/g(u)^2$, and
$\tilde{g}'''_\gamma(u)= \gamma \alpha^3\,
	\bigl(g'''(u)/g(u) - 3\,g'(u)\,g''(u)/g(u)^2 + 2\,g'(u)^3/g(u)^3\bigr)$.
Now the invariant $\Theta_{123\gamma}(g(u))$ is obtained by defining
a suitable ratio of the derivatives 
such that both $\gamma$ and $\alpha$ cancel out:
{\large
\begin{equation} \label{eq:Th123G}
 \begin{tabular}{ll}
   {\normalsize $\Theta_{123\gamma}(g(u))$} &=
        $\frac{\tilde{g}'_\gamma(u)\,\tilde{g}'''_\gamma(u)}
                        {\tilde{g}''_\gamma(u)^2}$ \\
        \rule[-3.0mm]{0mm}{10mm}
        &= $\frac{g^2 g'\,g''' - 3\,g\,{g'}^2 g'' + 2\,{g'}^4}
                 {g^2 {g''}^2  - 2\,g\,{g'}^2 g'' +  \ {g'}^4}$
 \end{tabular}
\end{equation}
}
Analogously to eq.~(\ref{eq:Thm12G}),
we can define a modified invariant
{\large
\begin{equation}
 \label{eq:Thm123G}
 \begin{tabular}{lcl}
       &0 &{\normalsize if cond1} \\  \rule[-4mm]{0mm}{8mm}
    {\normalsize $\Theta_{m123\gamma}=$}
       &$\frac{g^2 g'\,g''' - 3\,g\,{g'}^2 g'' + 2\,{g'}^4}
              {g^2 {g''}^2  - 2\,g\,{g'}^2 g'' +  \ {g'}^4}$
          &{\normalsize if cond2} \\
       &$\frac{g^2 {g''}^2  - 2\,g\,{g'}^2 g'' +  \ {g'}^4}
              {g^2 g'\,g''' - 3\,g\,{g'}^2 g'' + 2\,{g'}^4}$
          &{\normalsize else} \\
 \end{tabular}
\end{equation}
}
where
condition cond1 is
\mbox{$g^2 g'\,g''' - 3\,g\,{g'}^2 g'' + 2\,{g'}^4 = 0$} $\wedge$
$g^2 {g''}^2  - 2\,g\,{g'}^2 g'' +  \ {g'}^4 = 0$,
and condition cond2 is
$|{g^2 g'\,g''' - 3\,g\,{g'}^2 g'' + 2\,{g'}^4}|$ $<$
$|{g^2 {g''}^2  - 2\,g\,{g'}^2 g'' +  \ {g'}^4}|$.
Again, this modification entails
\mbox{$-1 \leq \Theta_{m123\gamma} \leq 1$}.

%--------------------------------------------------------------------------
\subsection{An Analytical Example}
%--------------------------------------------------------------------------

It is a straightforward albeit cumbersome exercise to verify
the invariants from eqs.~(\ref{eq:Th12G})
and~(\ref{eq:Th123G}) with an analytical, differentiable function.
As an arbitrary example, we choose 
\[
  f(x)=3 x\sin(2\pi x) + 30
\]
The first three derivatives are
$f'(x)=3 \sin(2\pi x) + 6\pi x \cos(2\pi x)$,
$f''(x)=12\pi \cos(2\pi x) - 12\pi^2 x \sin(2\pi x)$, and
$f'''(x)=-36\pi^2 \sin(2\pi x) - 24\pi^3 x \cos(2\pi x)$.
Then, according to eq.~(\ref{eq:Th12G}),
$
  \Theta_{12\gamma}(f(x))=
	  (3 x\sin(2\pi x) + 30)\,(3 x\sin(2\pi x) +
		6\pi x \cos(2\pi x)) \,/\, 
	\bigl((3 x\sin(2\pi x) + 30)\, (12\pi \cos(2\pi x) - 
		12\pi^2 x \sin(2\pi x))-
		(3 \sin(2\pi x) + 6\pi x \cos(2\pi x))^2\bigr)
$.

If we now replace $f(x)$ with a gamma corrected version, say
$f_{0.45}(x)=255^{1-0.45}\cdot3 x\sin(2\pi x) + 30)^{0.45}$,
the first derivative becomes
$f'_{0.45}(x)=255^{0.55}\cdot 0.45\, (3 \sin(2\pi x) + 30)^{-0.55}
	(3 \sin(2\pi x) + 6\pi x \cos(2\pi x))$,
the second derivative is
$f''_{0.45}(x)=-255^{0.55}\cdot 0.45\cdot 0.55\, (3 \sin(2\pi x) + 30)^{-1.55}
	(3 \sin(2\pi x) + 6\pi x \cos(2\pi x))^2 + 
	255^{0.55}\cdot 0.45\, (3 x\sin(2\pi x) + 30)^{-0.55}
	(12\pi \cos(2\pi x) - 12\pi^2 x \sin(2\pi x))$,
and the third is
$f'''_{0.45}(x)=255^{0.55}\cdot 0.45 \, (3 \sin(2\pi x) + 30)^{-0.55}
	\bigl(1.55\cdot 0.55\, (3 \sin(2\pi x) + 30)^{-2}
	(3 \sin(2\pi x) + 6\pi x \cos(2\pi x))^3 + 
	\ (-3)\cdot 0.55\, (3 \sin(2\pi x) + 30)^{-1}
	(3 \sin(2\pi x) + 6\pi x \cos(2\pi x))\, 
	(12\pi \cos(2\pi x) - 12\pi^2 x \sin(2\pi x)) +
	(-36\pi^2 \sin(2\pi x) - 24\pi^3 x \cos(2\pi x)) \bigr)$.
If we plug these derivatives into eq.~(\ref{eq:Th12G}), we obtain 
an expression for $\Theta_{12\gamma}(f_{0.45}(x))$ which is identical to the one
for $\Theta_{12\gamma}(f(x))$ above.
The algebraically inclined reader is encouraged to verify the invariant
$\Theta_{123\gamma}$ for the same function.

\begin{figure}[htb]
 \begin{center}
  \epsfig{file=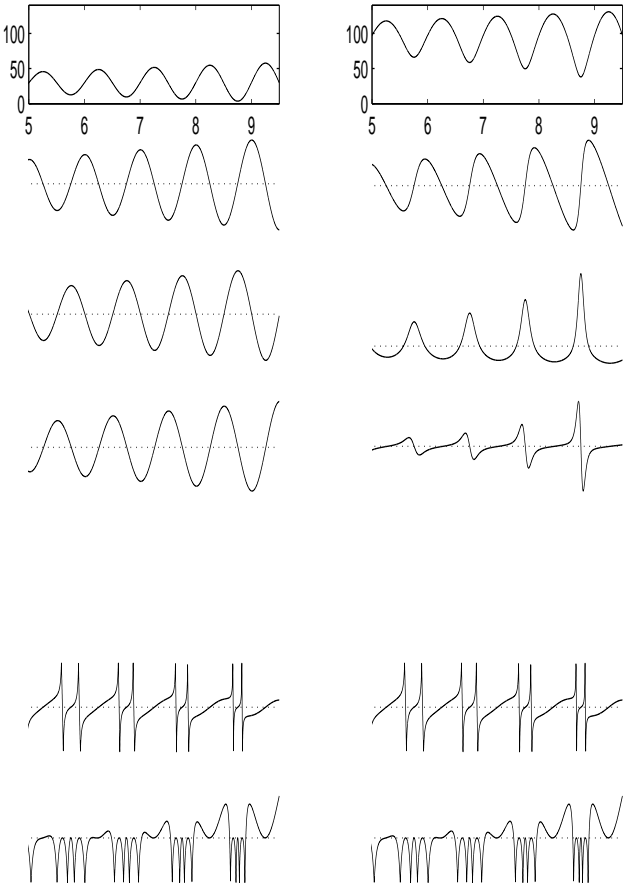, width=8.4cm,height=12cm}
  \caption
        {An analytical example function.
	\ (left) $f(x)=3 x\sin(2\pi x) + 30$;
	\ (right) $f_\gamma(x)=p\,f(x)^\gamma$, $\gamma=0.45$.
        \ (first~row)~original functions %, before and after gamma correction;
        \ (second~row)~first derivatives;
        \ (third~row)~second derivatives;
        \ (fourth~row)~third derivatives;
	\ (fifth~row)~$\Theta_{m12\gamma}$;
	\ (sixth~row)~$\Theta_{m123\gamma}$.
        }
  \label{fig:AnalyEx}
 \end{center}
\end{figure}
Fig.~\ref{fig:AnalyEx} shows the example function and its 
gamma corrected counterpart, together with their derivatives
and the two modified invariants.
As expected, the graphs of the invariants are the same on the right as
on the left.
Note that the invariants define a many-to-one mapping.
That is, the mapping is not information preserving,
and it is not possible to reconstruct the original image from its
invariant representation.

%--------------------------------------------------------------------------
\subsection{Extension to 2-d}
%--------------------------------------------------------------------------

If $\Theta_{m12\gamma}$ or $\Theta_{m123\gamma}$ are to be computed on images,
then eqs.~(\ref{eq:Th12G}) to~(\ref{eq:Thm123G})
have to be generalized to two dimensions.
This is to be done in a rotationally invariant way
in order to achieve invariance under similarity transformations.
The standard way is to use rotationally symmetric operators.
For the first derivative, we have the well known {\em gradient magnitude},
defined as
\begin{equation}
 \label{eq:gm}
   \nabla(x,y)=\sqrt{I_x^2+I_y^2} := I'
\end{equation}
where $I(x,y)$ is the 2-d image function, and $I_x$, $I_y$ are partial
derivatives along the x-axis and the y-axis.
For the second order derivative, we can use
the linear {\em Laplacian}
\begin{equation}
 \label{eq:lapl}
   \nabla^2(x,y)=I_{xx}+I_{yy} :=I''
\end{equation}
Horn~\cite{horn86} also presents an alternative second order derivative
operator, the {\em quadratic variation}\footnote{Actually, unlike Horn,
we have taken the square root.}
\begin{equation}
 \label{eq:qv}
   \mathrm{QV}(x,y)=\sqrt{I_{xx}^2 + 2 I_{xy}^2 + I_{yy}^2}
\end{equation}
Since the QV is not a linear operator and more expensive to compute,
we use the Laplacian for our implementation.
For the third order derivative, we can define,
in close analogy with the quadratic variation,
a {\em cubic variation} as
\begin{equation}
 \label{eq:cv}
   \mathrm{CV}(x,y)=\sqrt{I_{xxx}^2 + 3 I_{xxy}^2 + 3 I_{xyy}^2 + I_{yyy}^2}
	:= I'''
\end{equation}

The invariants from eqs.~(\ref{eq:Th12G}) to~(\ref{eq:Thm123G})
remain valid in 2-d if we replace $f'$ with $I'$, $f''$ with $I''$, and
$f'''$ with $I'''$.
This can be verified by going through the same argument as for the
\mbox{1-d} functions.
Recall that the critical observation in eq.~(\ref{eq:Th12G}) was that 
$\gamma$ cancels out, which is the case when all derivatives return
a factor $\gamma$. 
But such is also the case with the rotationally symmetric operators
mentioned above.
For example, if we apply the gradient magnitude operator to $\tilde{I}(x,y)$,
i.e.~to the logarithm of a gamma corrected image function,
we obtain
\[
  \nabla = \sqrt{\tilde{I}_x^2 + \tilde{I}_y^2}
		= \sqrt{\bigl(\gamma\,\frac{I_x}{I}\bigr)^2 +
			\bigl(\gamma\,\frac{I_y}{I}\bigr)^2}
		= \gamma\,\frac{\sqrt{I_x^2 + I_y^2}}{I}
\]
returning a factor $\gamma$, and analogously for $\nabla^2$, QV, and CV.
A similar argument holds for eq.~(\ref{eq:Th123G}) where we have to show,
in addition, that the first derivative returns a factor $\alpha$,
the second derivative returns a factor $\alpha^2$, 
and the third derivative returns a factor $\alpha^3$,
which is the case for our 2-d operators.

%--------------------------------------------------------------------------
\subsection{Differential Operators}
%--------------------------------------------------------------------------

\begin{figure}[htb]
 \begin{center}
  \epsfig{file=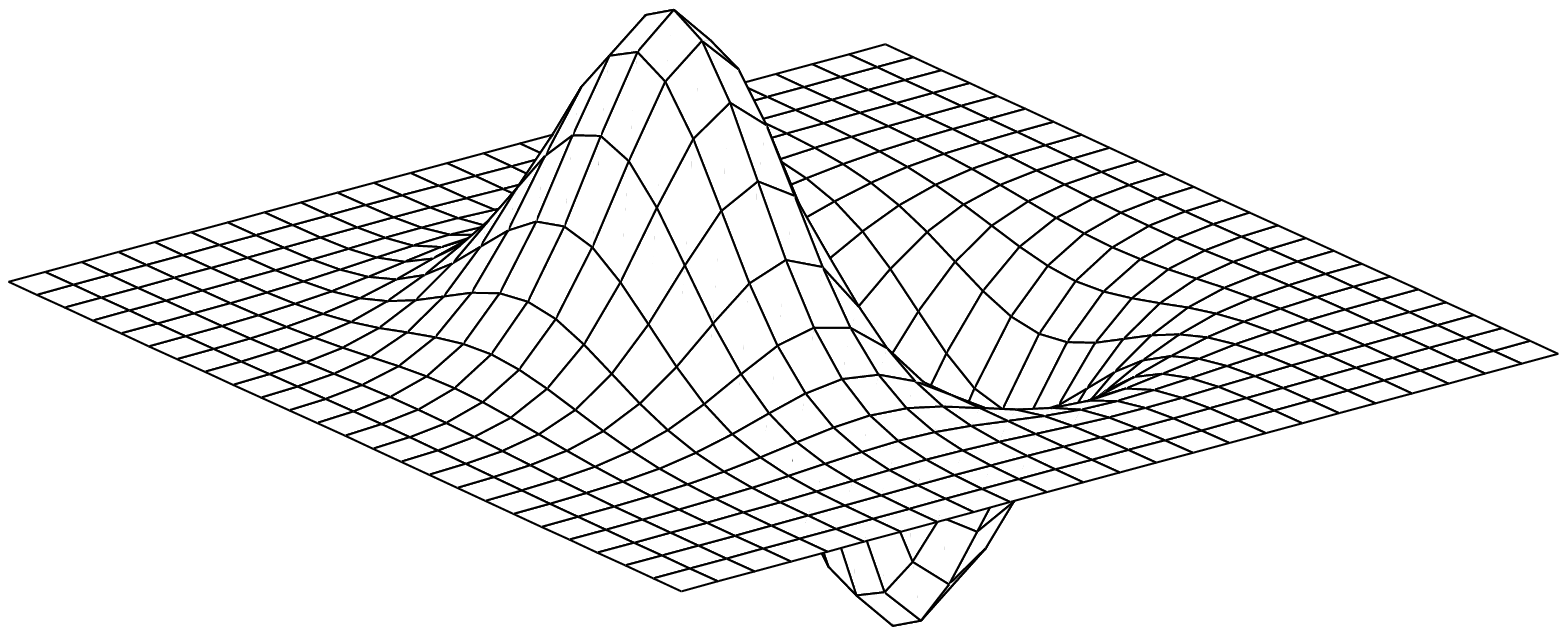,  width=2.7cm,height=3.3cm}
  \epsfig{file=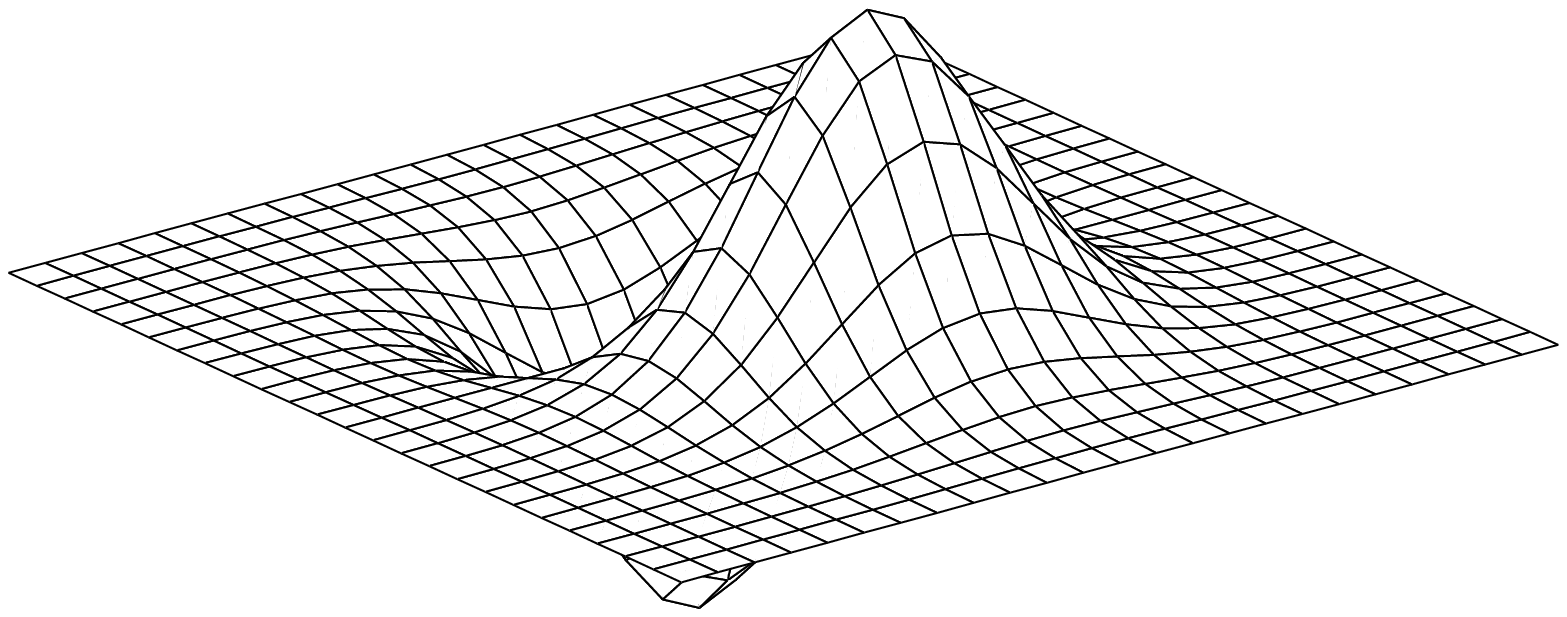,  width=2.7cm,height=3.3cm}
  \epsfig{file=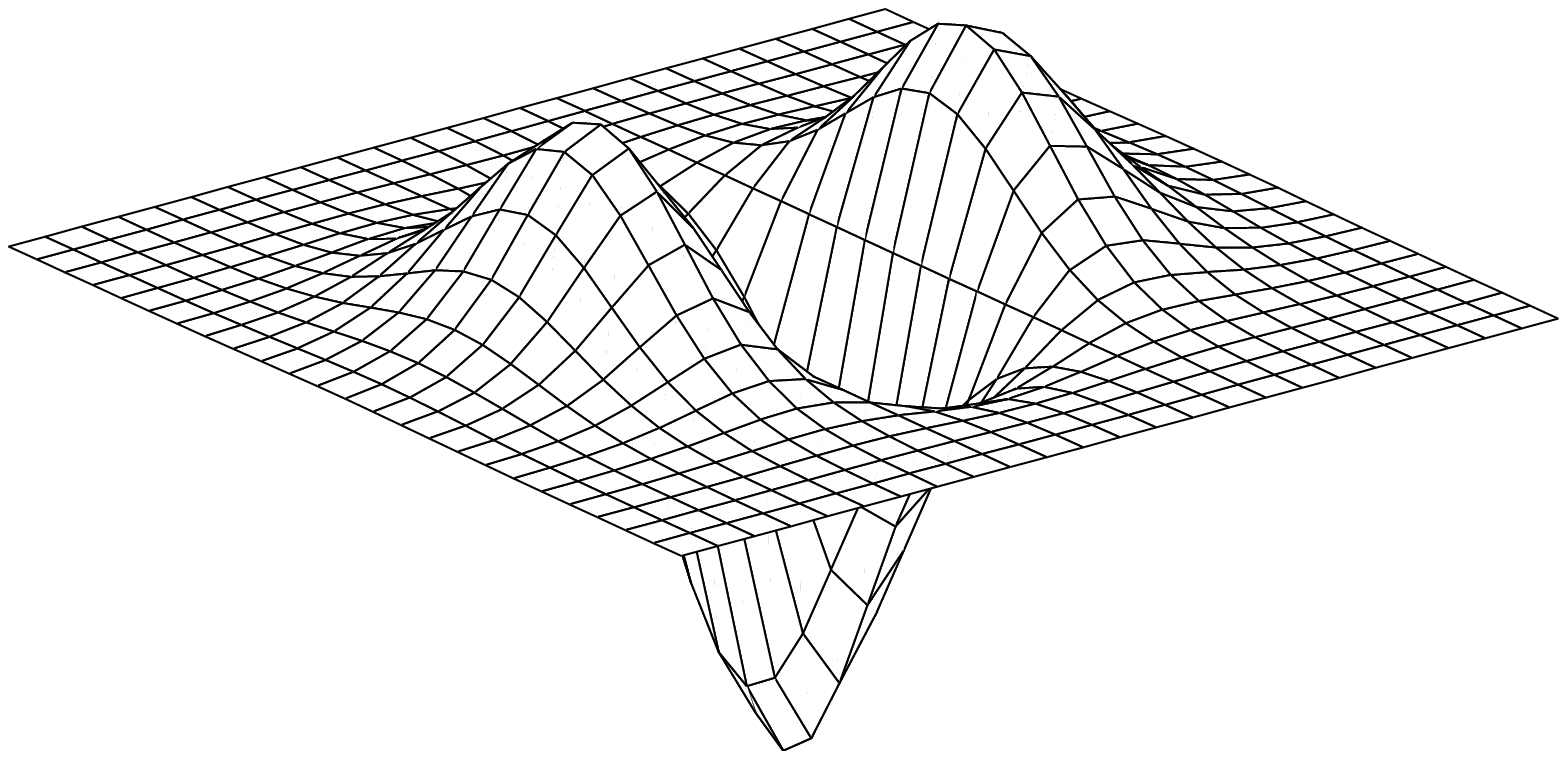, width=2.7cm,height=3.3cm}
  \epsfig{file=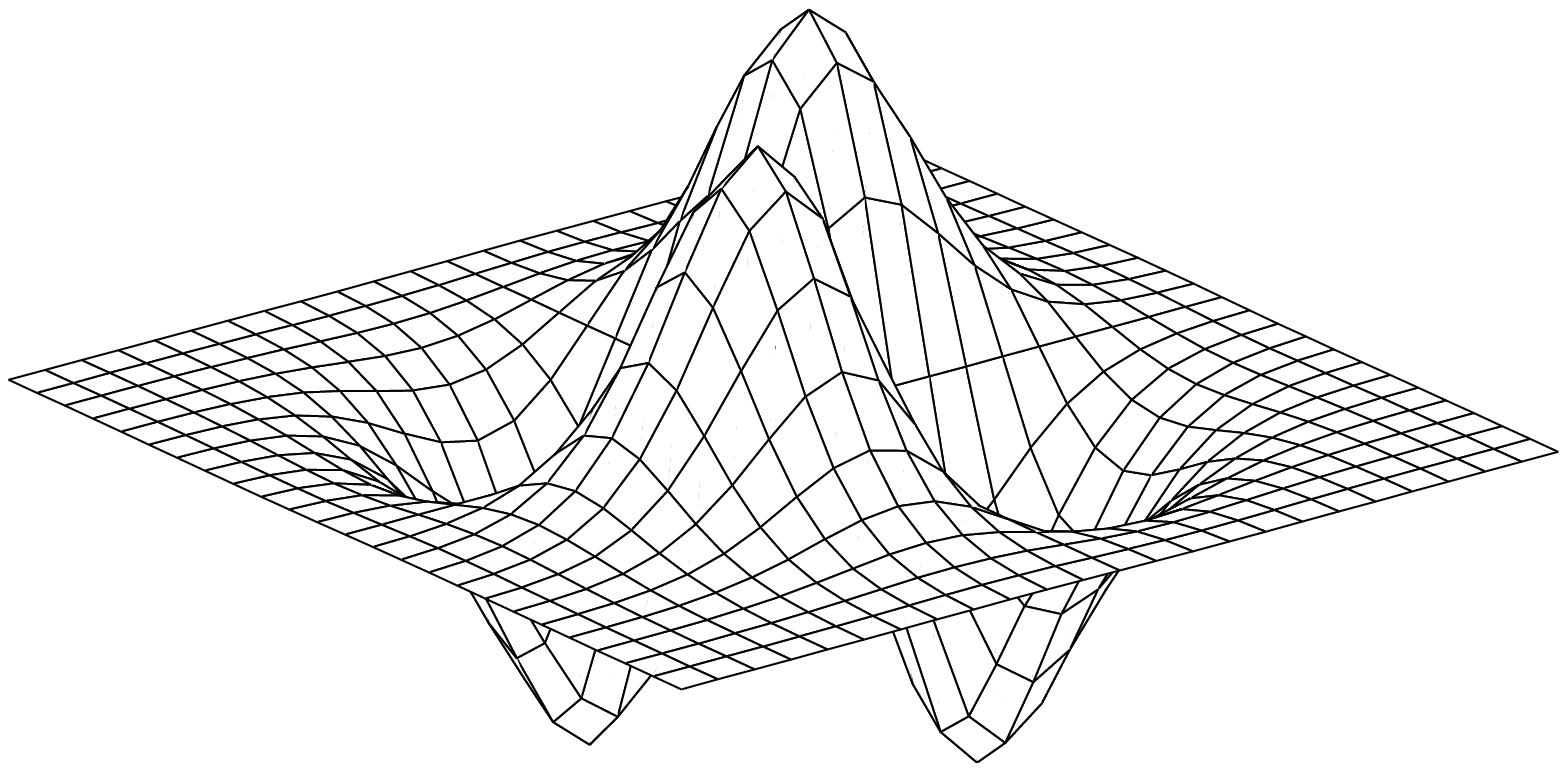, width=2.7cm,height=3.3cm}
  \epsfig{file=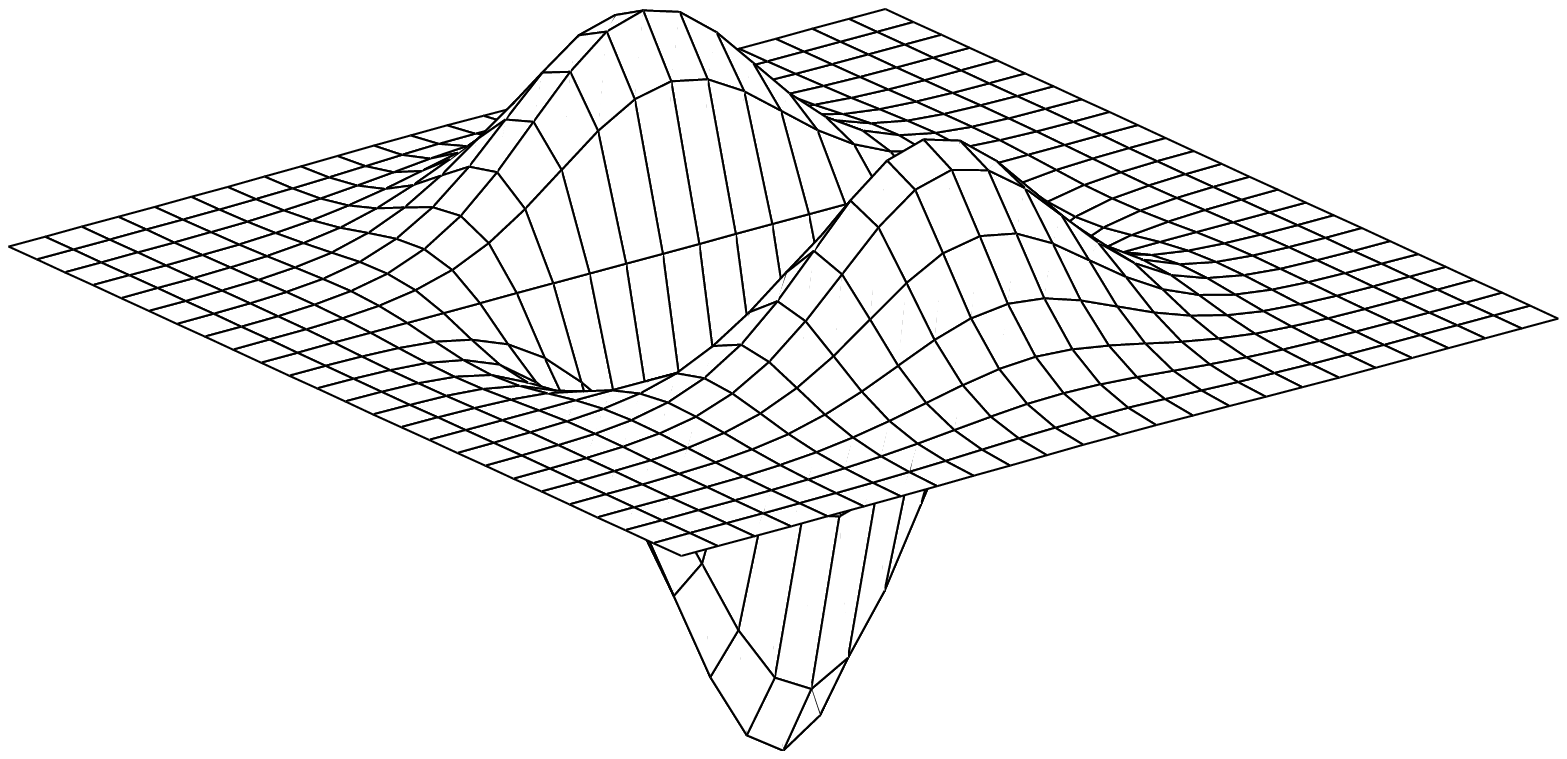, width=2.7cm,height=3.3cm}
  \epsfig{file=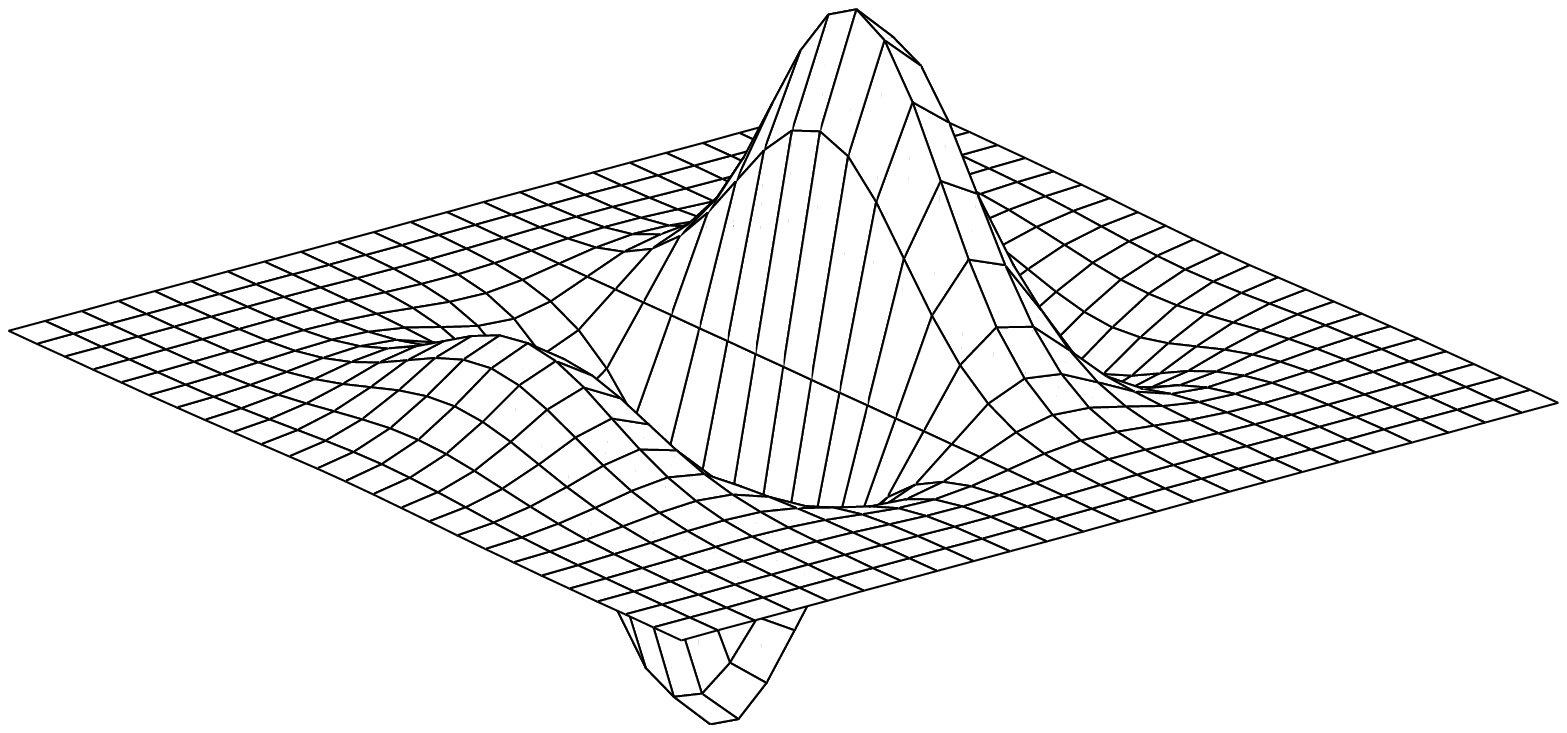,width=2.7cm,height=3.3cm}
  \epsfig{file=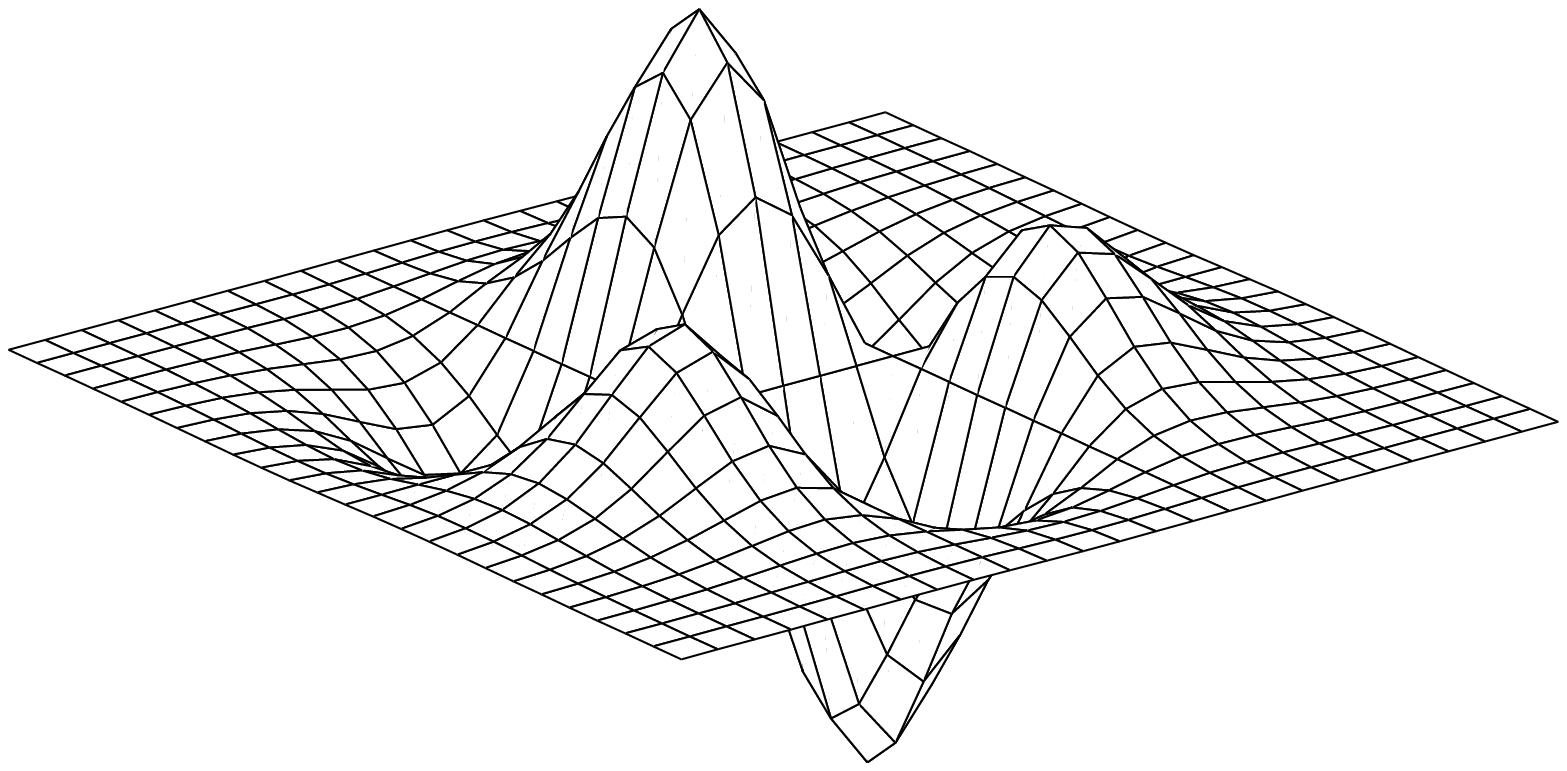,width=2.7cm,height=3.3cm}
  \epsfig{file=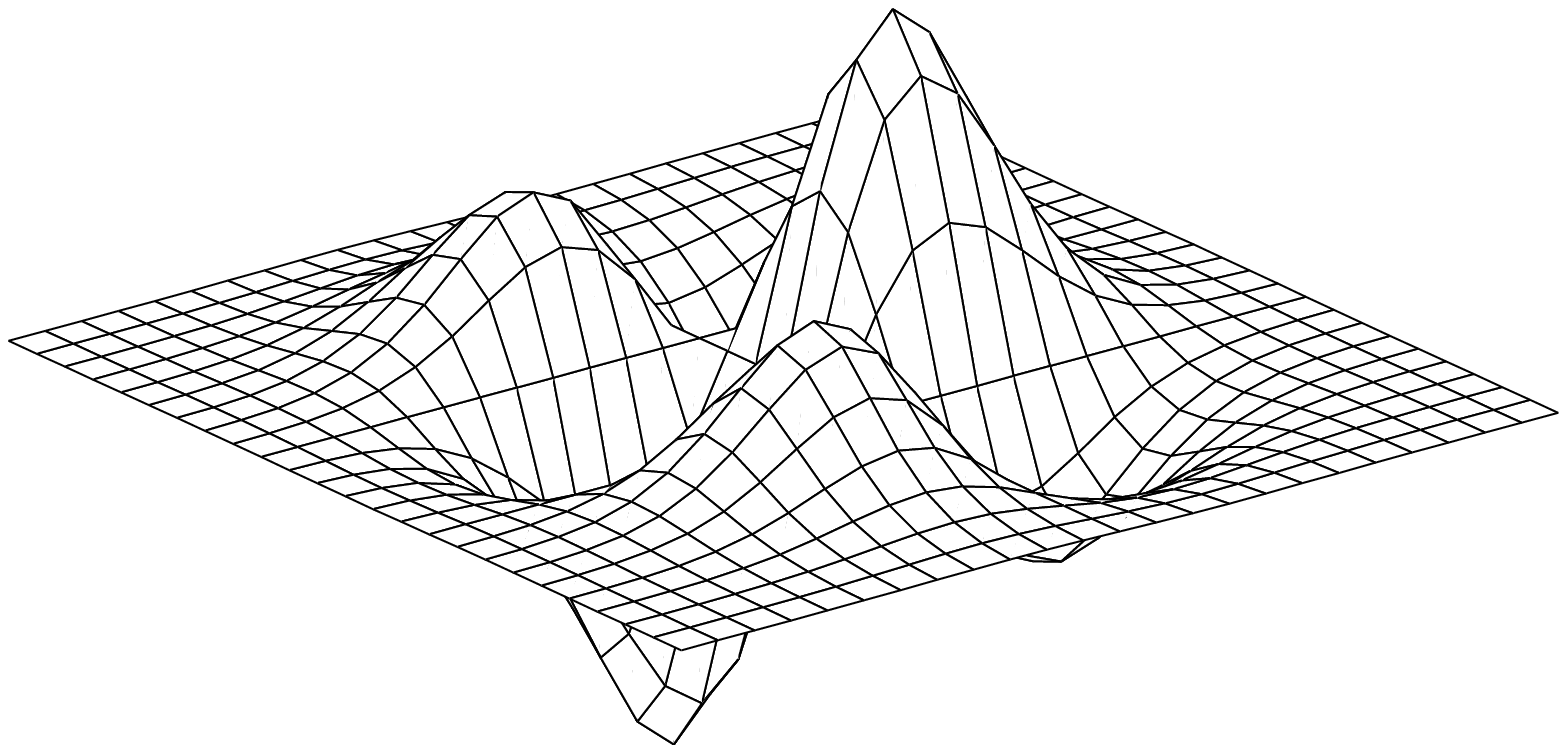,width=2.7cm,height=3.3cm}
  \epsfig{file=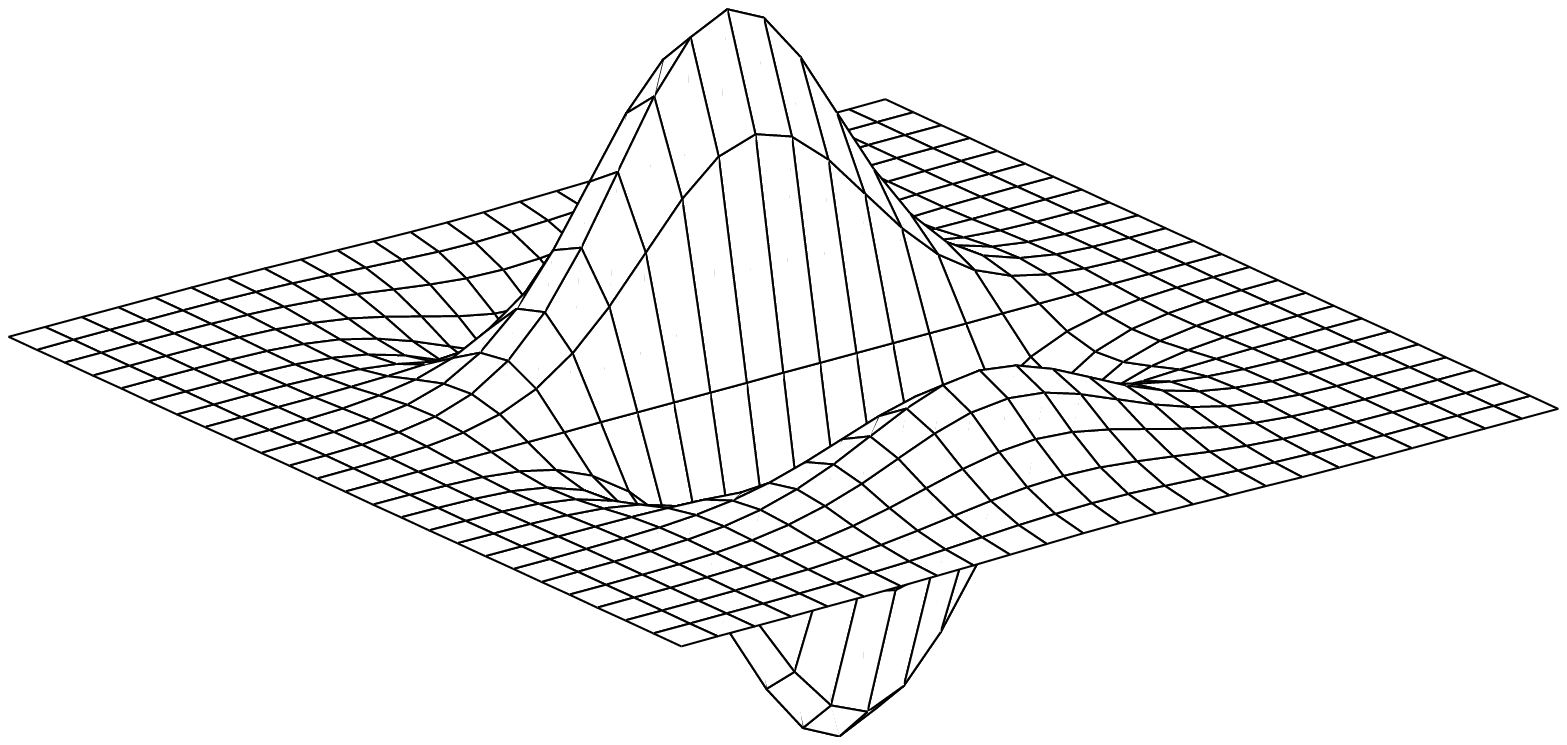,width=2.7cm,height=3.3cm}
  \caption
        {Partial derivatives of the 2-d Gaussian.
        \ (first row) $G_x$, $G_y$, $G_{xx}$;
        \ (second row) $G_{xy}$, $G_{yy}$, $G_{xxx}$;
        \ (third row) $G_{xxy}$, $G_{xyy}$, $G_{yyy}$.
        }
  \label{fig:GaussKernels}
 \end{center}
\end{figure}

While the derivatives of continuous, differentiable functions
are uniquely defined,
there are many ways to implement derivatives for {\em sampled} functions.
We follow Schmid and Mohr~\cite{sm97}, ter Haar Romeny~\cite{r94},
and many other researchers in employing the derivatives
of the Gaussian function as filters
to compute the derivatives of a sampled image function via convolution.
This way, derivation is combined with smoothing.
The 2-d zero mean Gaussian is defined as
\begin{equation} 
 \label{eq:Gauss2d}
  G = \frac{1}{2 \pi\,\sigma^2}\ e^{-\frac{x^2+y^2}{2 \sigma^2}}
\end{equation}
The partial derivatives up to third order are
$G_x=-x/\sigma^2\,G$,
$G_y=-y/\sigma^2\,G$,
$G_{xx}=(x^2-\sigma^2)/\sigma^4\,G$,
$G_{xy}=xy/\sigma^4\,G$,
$G_{yy}=(y^2-\sigma^2)/\sigma^4\,G$,
$G_{xxx}=(3 \sigma^2 x-x^3)/\sigma^6\,G$,
$G_{xxy}=(\sigma^2 y-x^2y)/\sigma^6\,G$,
$G_{xyy}=(\sigma^2 x-xy^2)/\sigma^6\,G$,
$G_{yyy}=(3 \sigma^2 y-y^3)/\sigma^6\,G$.
They are shown in fig.~\ref{fig:GaussKernels}.
We used the parameter setting $\sigma=1.0$ and kernel size $7 \times 7$.
With these kernels, eq.~(\ref{eq:Th12G}), for example, is implemented as
{\small
\[
 \label{eq:Th12Gconv}
  \Theta_{12\gamma} =
	\frac{I \,\sqrt{(I \circledast G_x)^2 + (I \circledast G_y)^2}}
	     {I \,(I \circledast G_{xx} + I \circledast G_{yy}) \,-\,
		((I \circledast G_x)^2 + (I \circledast G_y)^2)}
\]
}
at each pixel $(x,y)$,
where $\circledast$ denotes convolution.

%****************************************************************************
\section{Experimental Data and Results}
%****************************************************************************

We evaluate the invariant $\Theta_{m12\gamma}$
from eq.~(\ref{eq:Thm12G}) in two different ways.
First, we measure how much the invariant computed on an image
without gamma correction is different from the invariant computed
on the same image but with gamma correction.
Theoretical, this difference should be zero, but in practice, it is not.
Second, we compare template matching accuracy on intensity images,
again without and with gamma correction, to the accuracy achievable if
instead the invariant representation is used.
We also examine whether the results can be improved by prefiltering.

%--------------------------------------------------------------------------
\subsection{Absolute and Relative Errors}
%--------------------------------------------------------------------------

A straightforward error measure is the {\em absolute error},
\begin{equation}
  \label{eq:AbsErr}
  \Delta_{GC}(i,j)=|\Theta_{GC}(i,j) - \Theta_{0GC}(i,j)|
\end{equation}
where "0GC" refers to the image without gamma correction, and GC stands
for either "SGC" if the gamma correction is done synthetically
via eq.~(\ref{eq:GammaCorr}),
or for "CGC" if the gamma correction is done via the camera hardware.
Like the invariant itself, the absolute error is computed
at each pixel location $(i,j)$ of the image,
except for the image boundaries
where the derivatives and therefore the invariants
cannot be computed reliably.

\begin{figure}[htbp]
 \begin{center}
  \epsfig{file=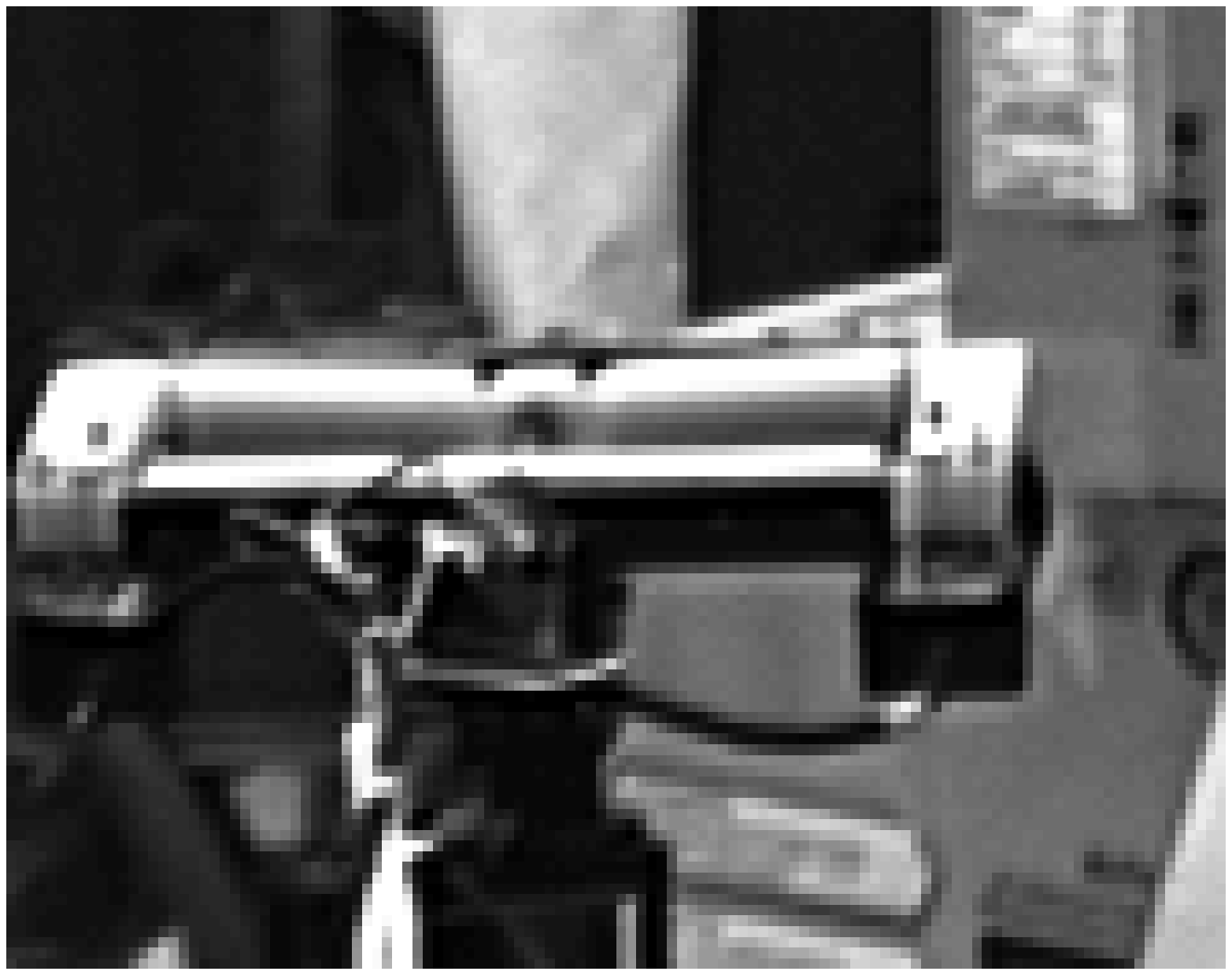,   width=2.7cm,height=2.4cm}
  \epsfig{file=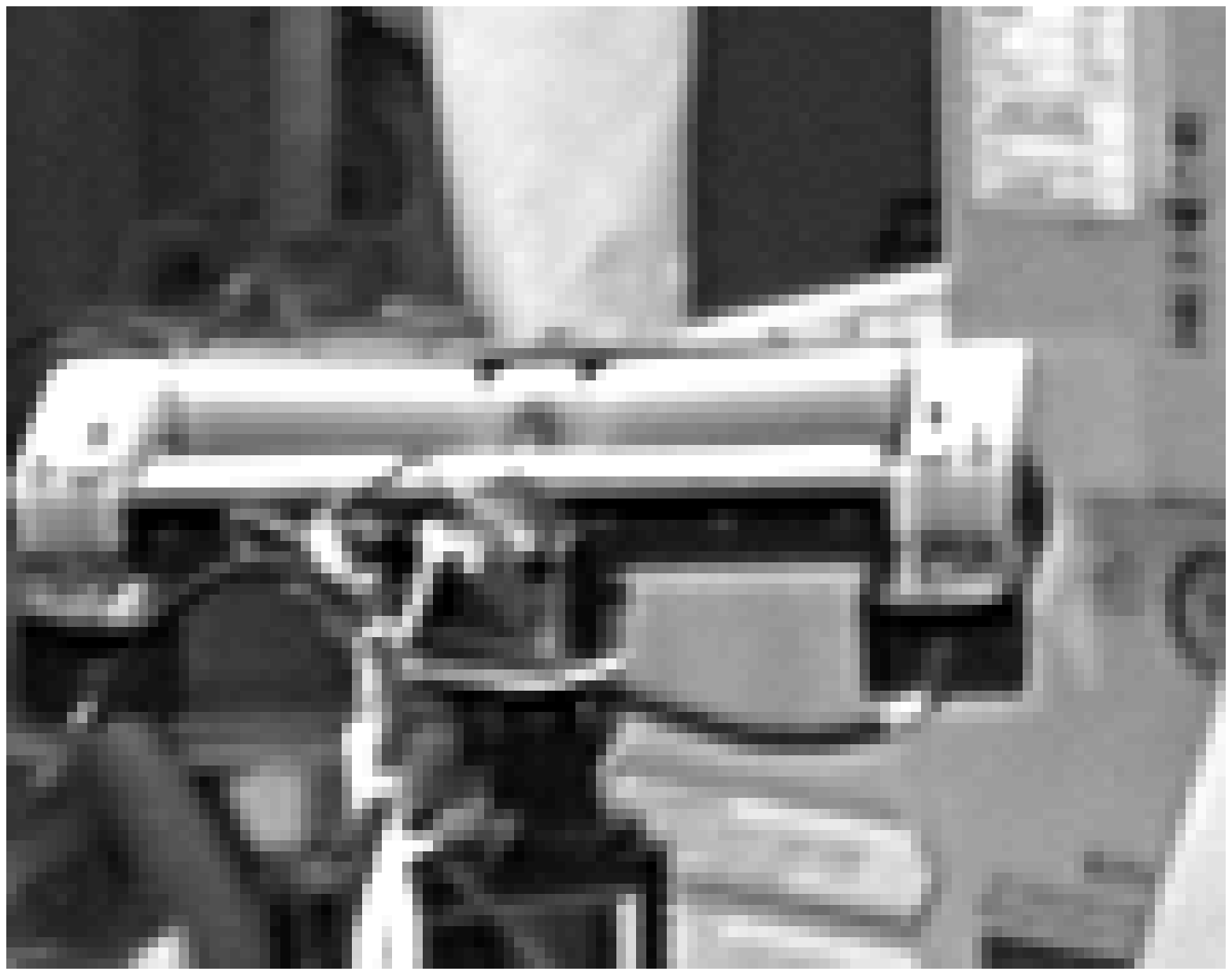,    width=2.7cm,height=2.4cm}
  \epsfig{file=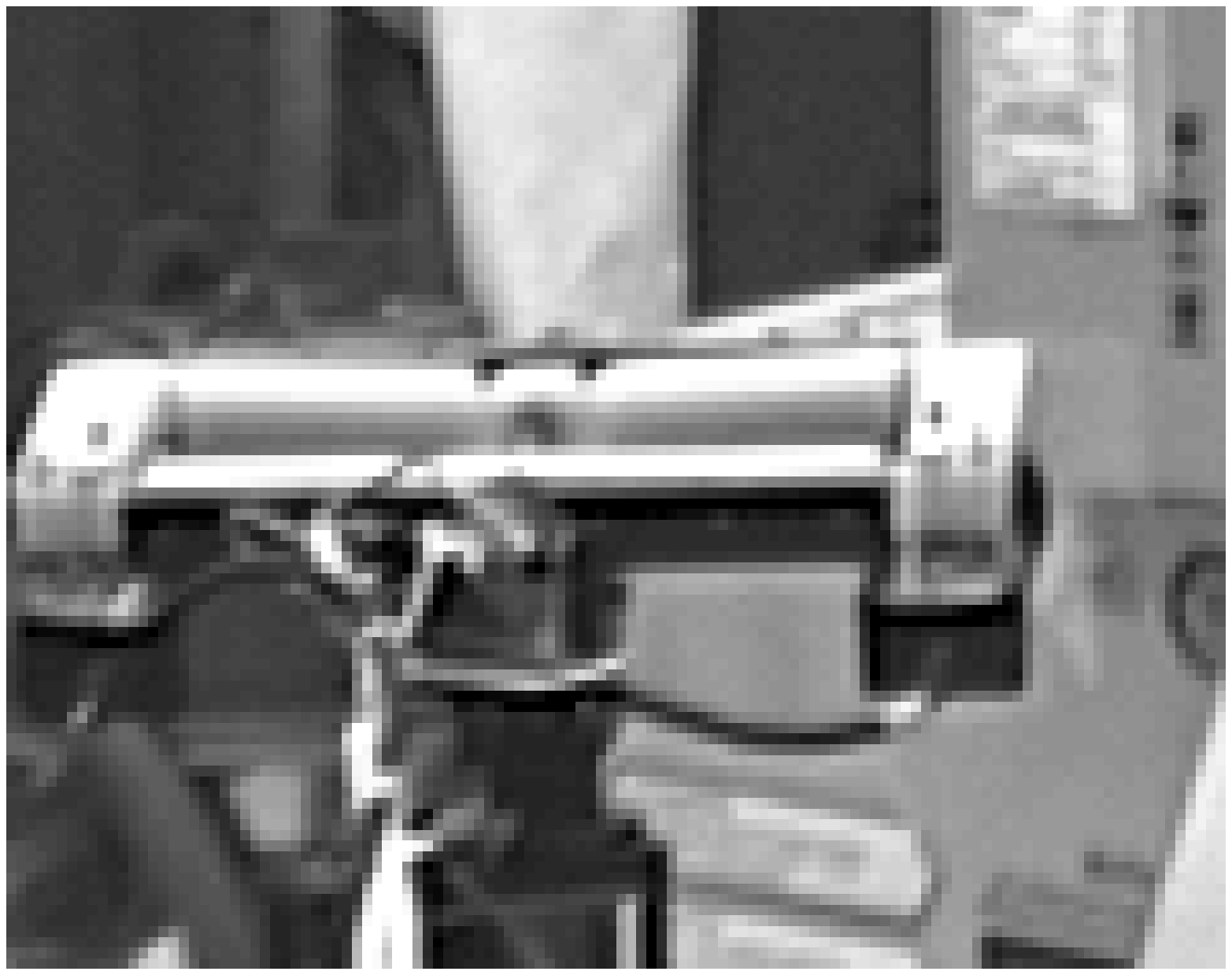,width=2.7cm,height=2.4cm}
  \caption
        {Example image {\tt WoBA}:
         \ (a)~no gamma correction, ``0GC'';
         \ (b)~gamma correction by camera, ``CGC'';
         \ (c)~synthetic gamma correction, ``SGC''.
         %  $\gamma$=0.60.
        }
  \label{fig:imas}
 \end{center}
\end{figure}

\begin{figure}[htbp]
 \begin{center}
  \subfigure[]{\epsfig{file=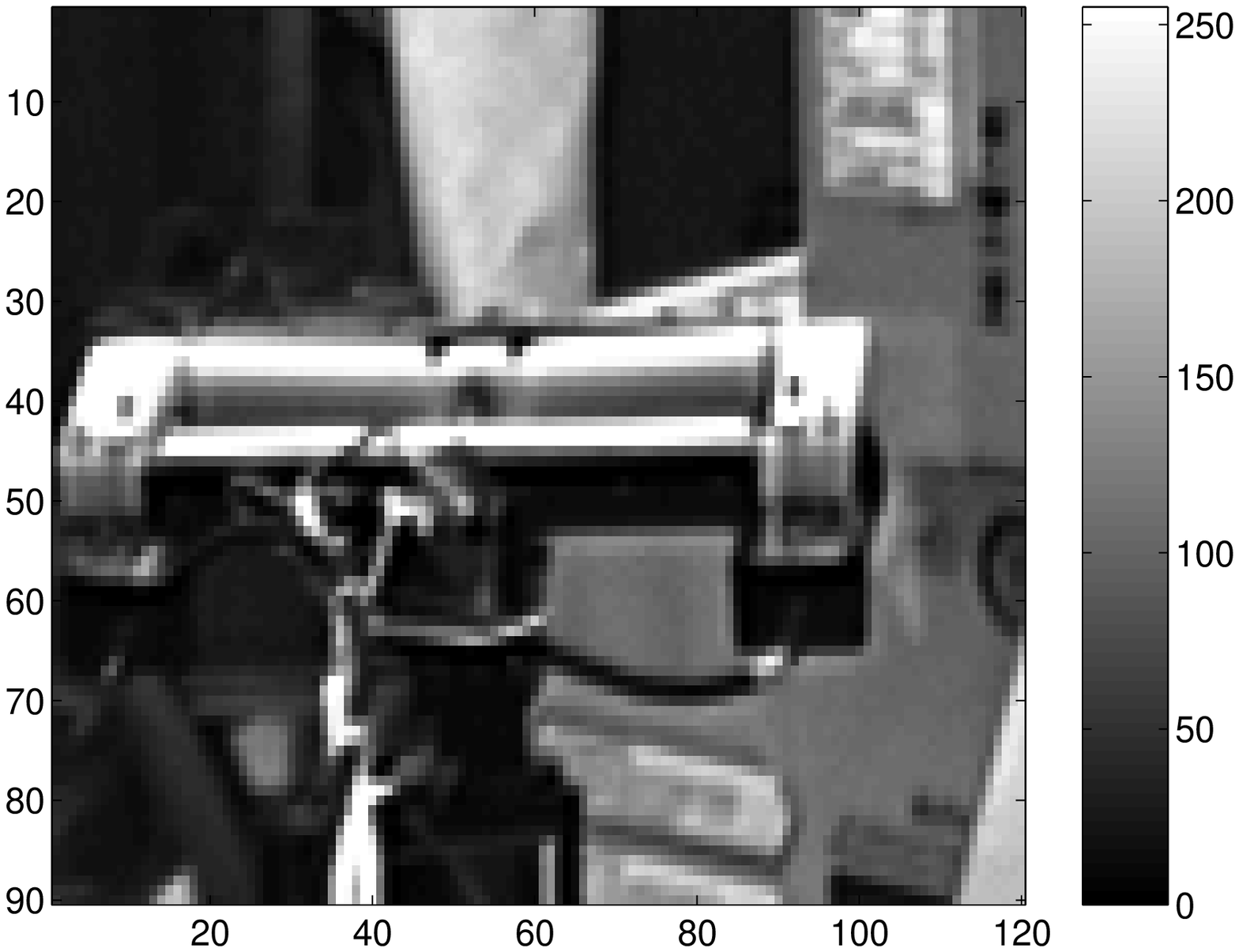,                     width=4.1cm}}
  \subfigure[]{\epsfig{file=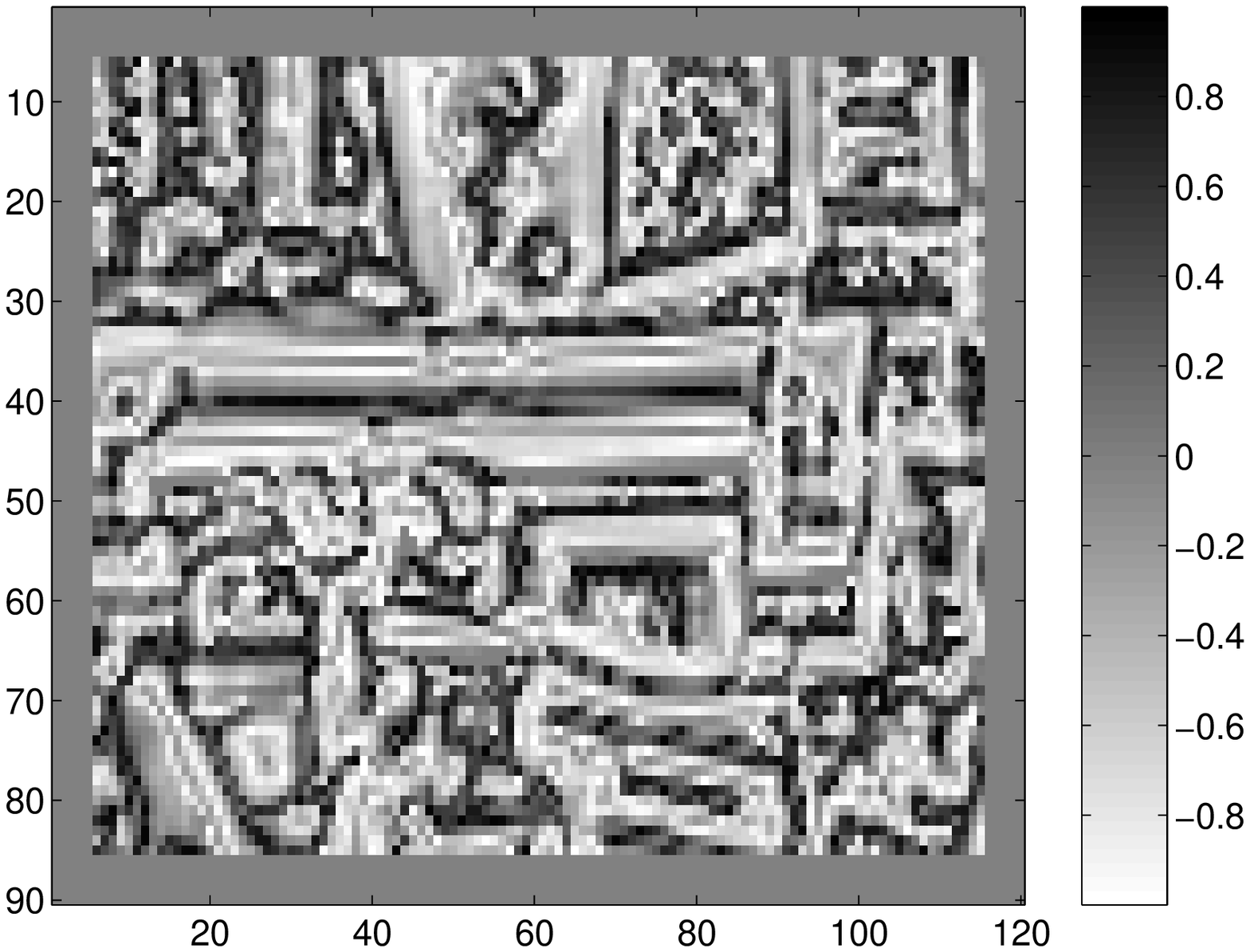,           width=4.1cm}}
  \subfigure[]{\epsfig{file=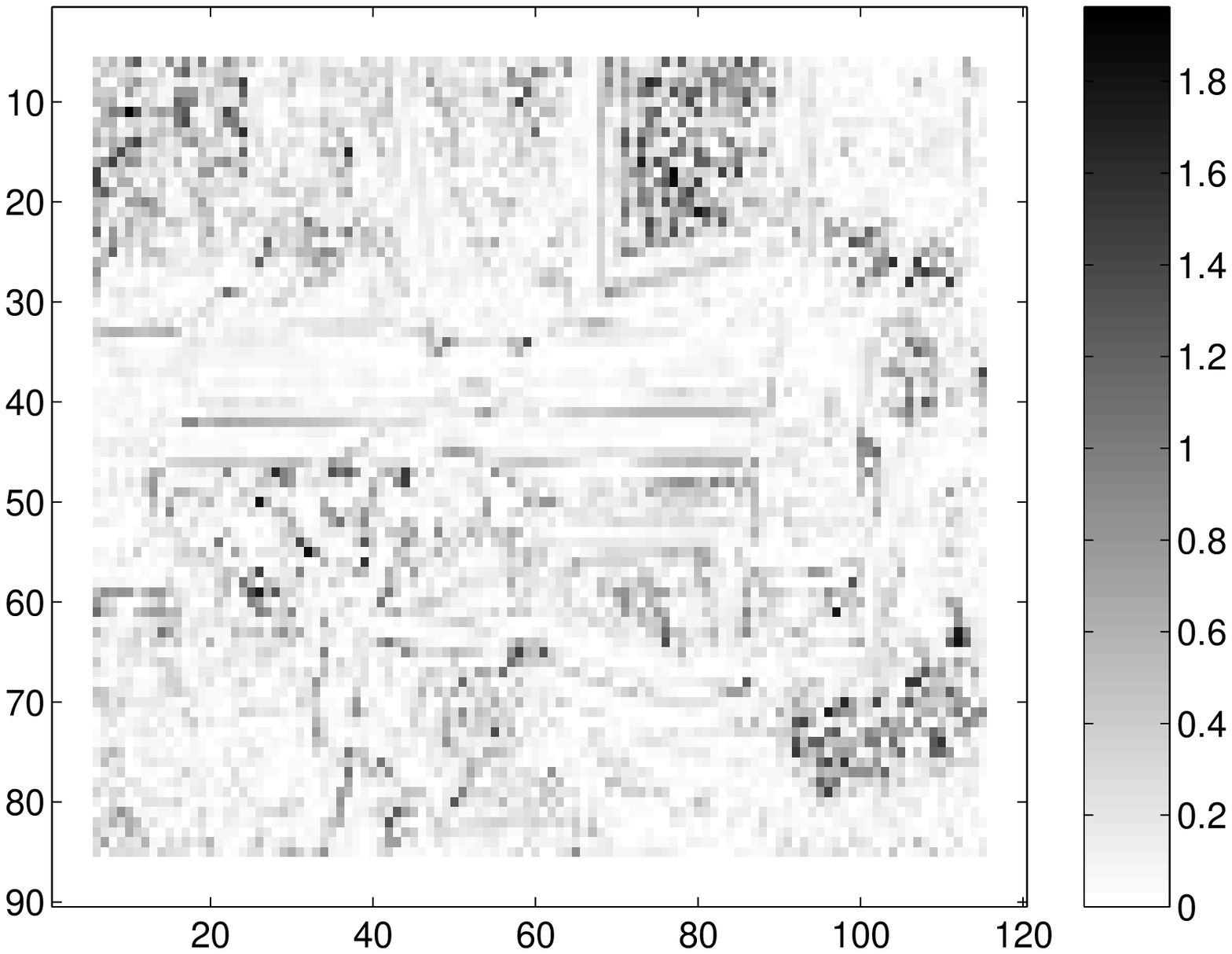,    width=4.1cm}}
  \subfigure[]{\epsfig{file=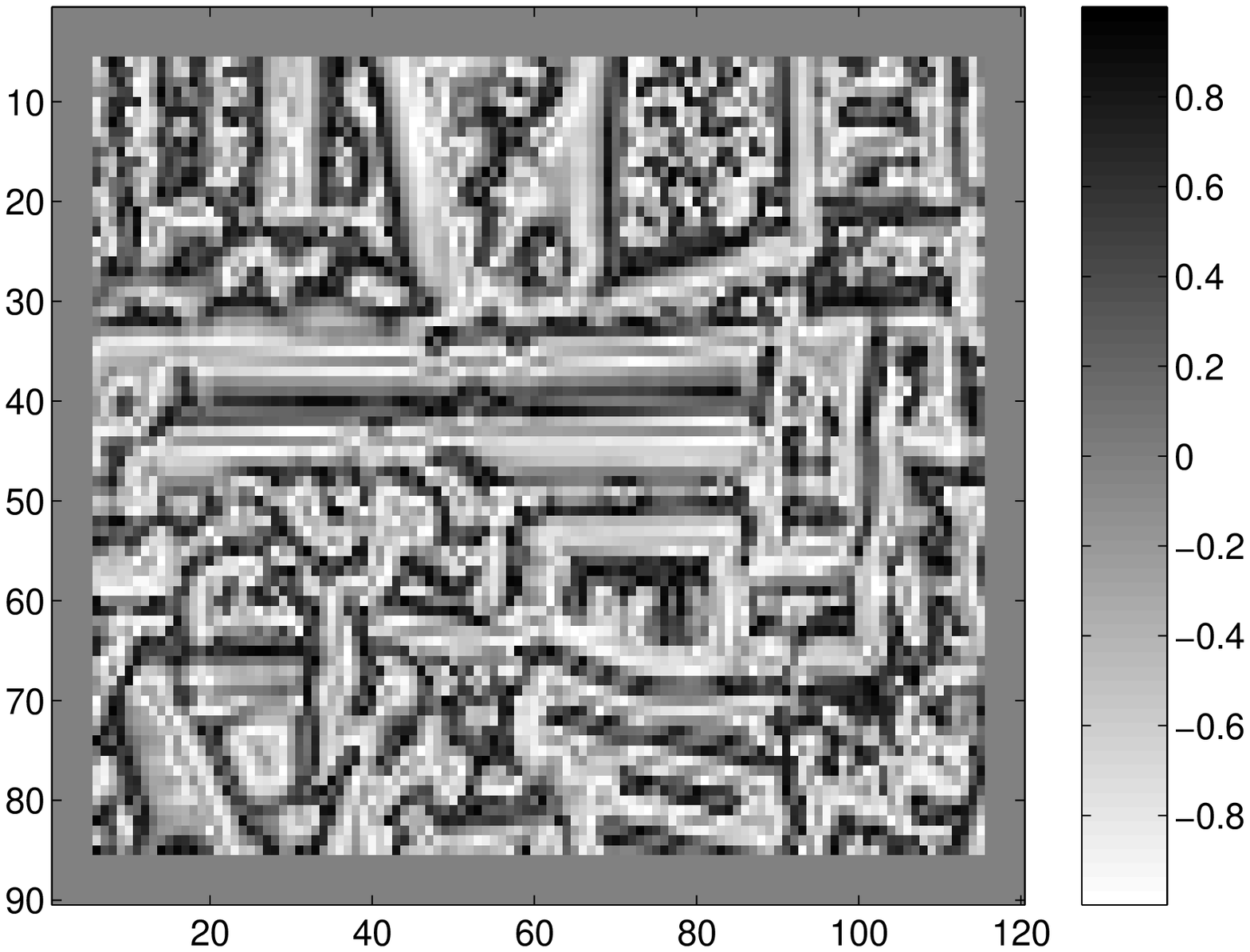,         width=4.1cm}}
  \subfigure[]{\epsfig{file=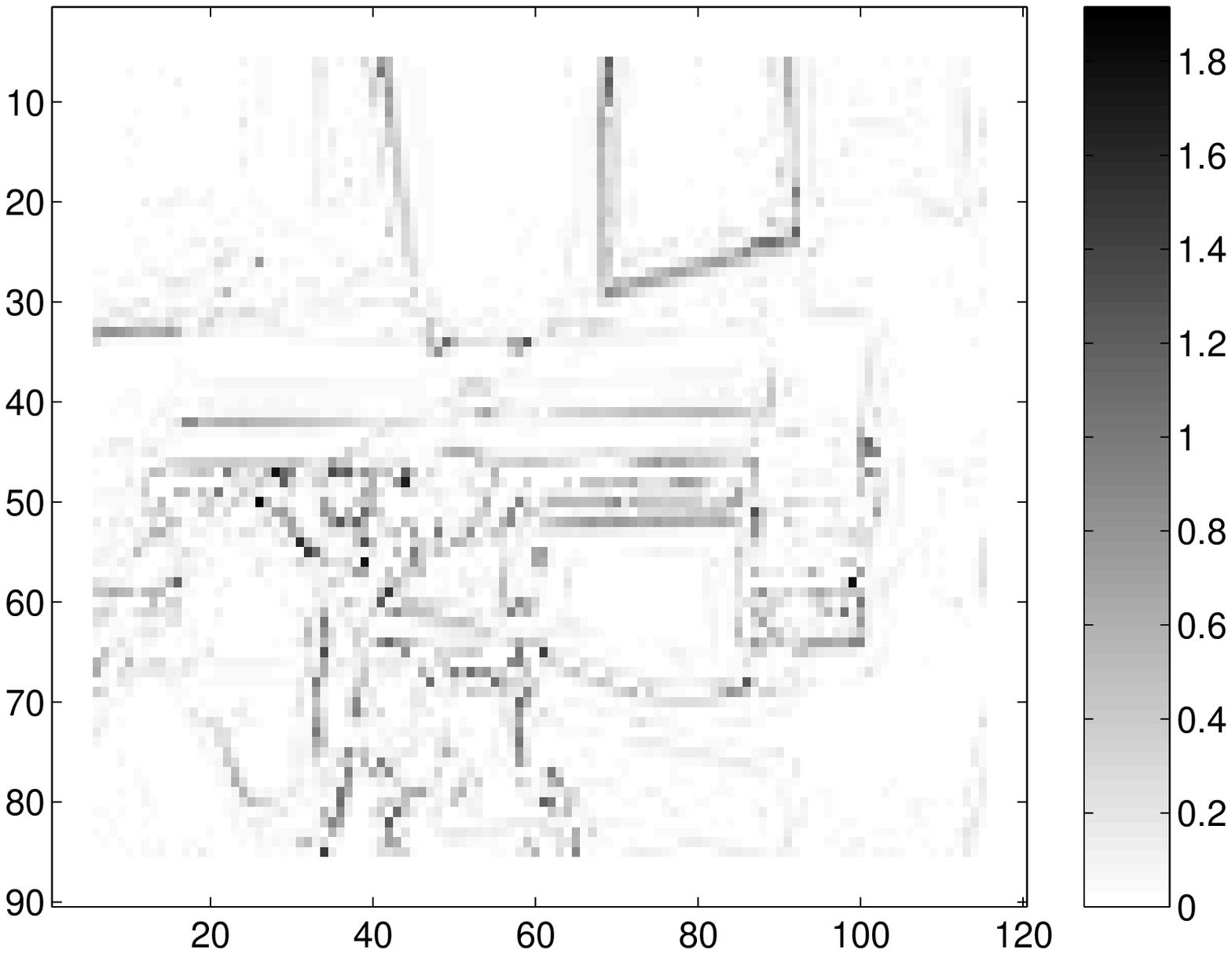,width=4.1cm}}
  \subfigure[]{\epsfig{file=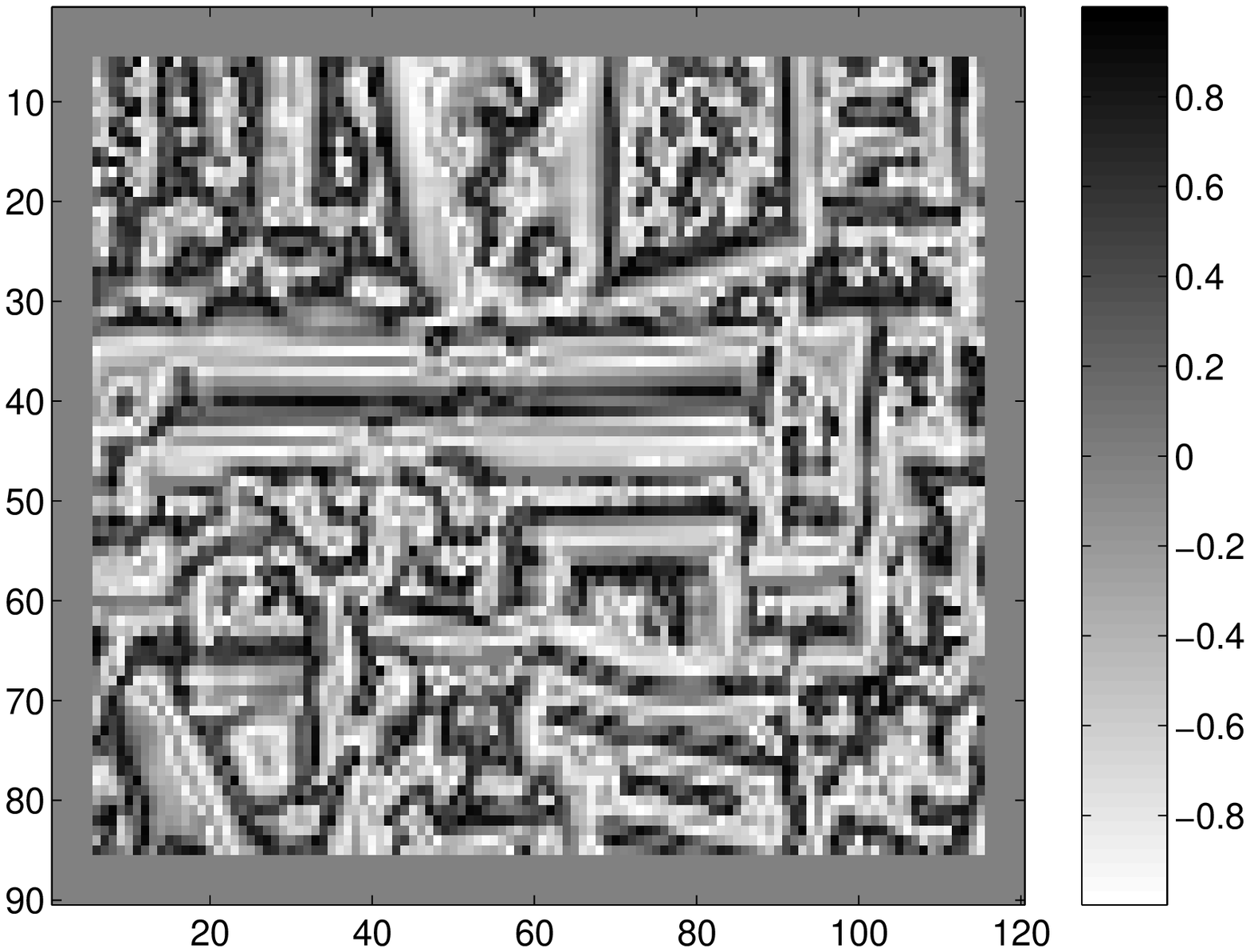,     width=4.1cm}}
  \caption
        {Absolute errors for invariant $\Theta_{m12\gamma}$, no prefiltering.
	(a)~image {\tt WoBA}, 0GC;
        (b)~$\Theta_{0GC}$;
        (c)~$\Delta_{CGC}$;
        (d)~$\Theta_{CGC}$;
        (e)~$\Delta_{SGC}$;
        (f)~$\Theta_{SGC}$.
        }
  \label{fig:AccuInv}
 \end{center}
\end{figure}

Fig.~\ref{fig:imas} shows an example image.
The SGC image has been computed from the 0GC image, with $\gamma=0.6$.
Note that the gamma correction is done {\em after}
the quantization of the 0GC image, since we don't have access to the
0GC image before quantization.

Fig.~\ref{fig:AccuInv} shows the invariant representations of the image data
from fig.~\ref{fig:imas} and the corresponding absolute errors.
Since \mbox{$-1 \leq \Theta_{m12\gamma} \leq 1$}, we have
\mbox{$0 \leq \Delta_{GC} \leq 2$}. 
The dark points in fig.~\ref{fig:AccuInv}, (c) and~(e),
indicate areas of large errors.
We observe two error sources:
\begin{itemize}
  \item The invariant cannot be computed robustly in homogeneous regions.
	This is hardly surprising, given that it is based on differentials
	which are by definition only sensitive to spatial changes of the signal.
  \item There are outliers even in the SGC invariant representation, at
	points of very high contrast edges.
	They are a byproduct of the inherent smoothing
	when the derivatives are computed with differentials of the Gaussian.
	Note that the latter put a ceiling on the maximum gradient magnitude
	that is computable on 8-bit images.
\end{itemize}

\begin{figure}[htbp]
 \begin{center}
  \subfigure[]{\epsfig{file=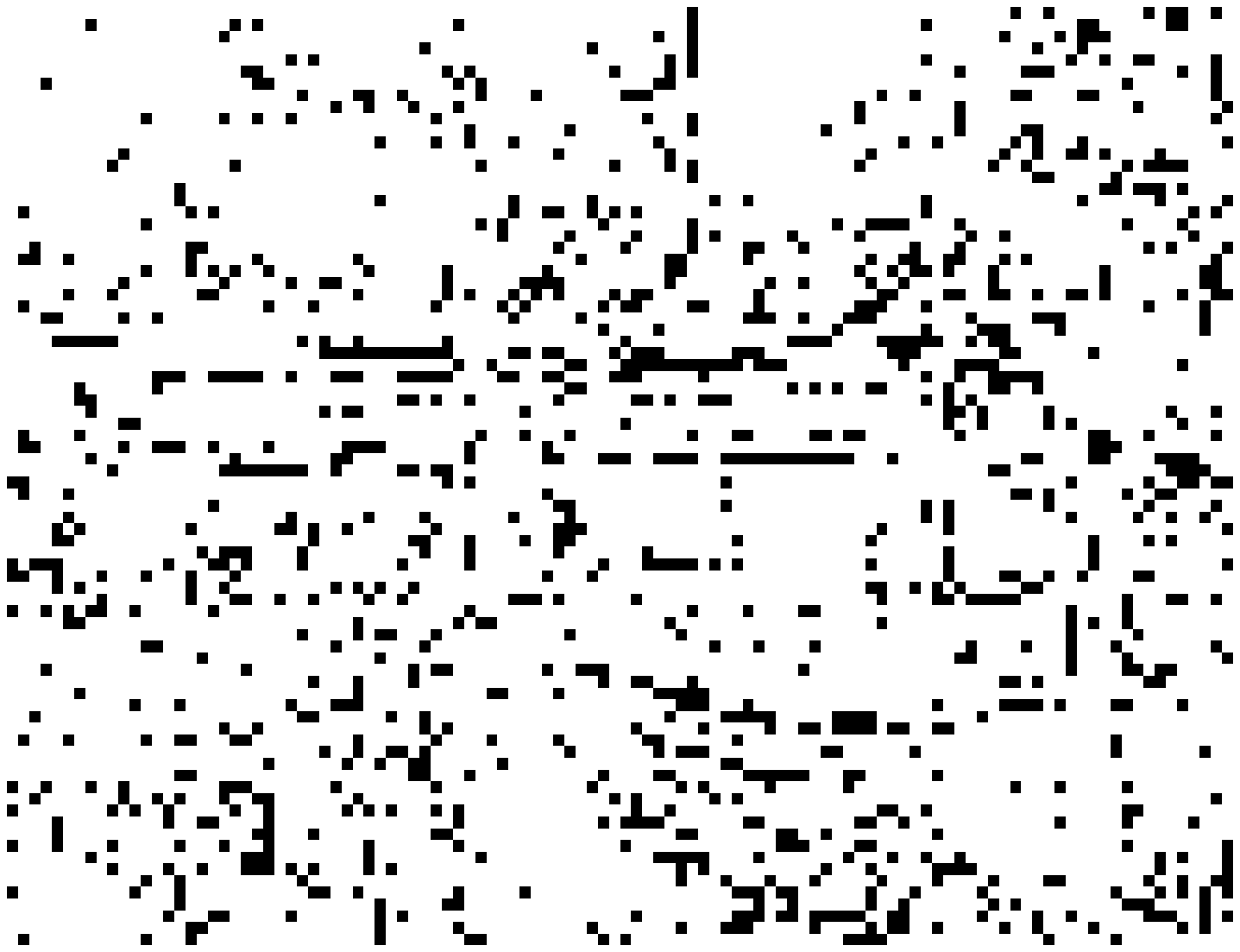,    width=2.7cm,height=2.5cm}}
  \subfigure[]{\epsfig{file=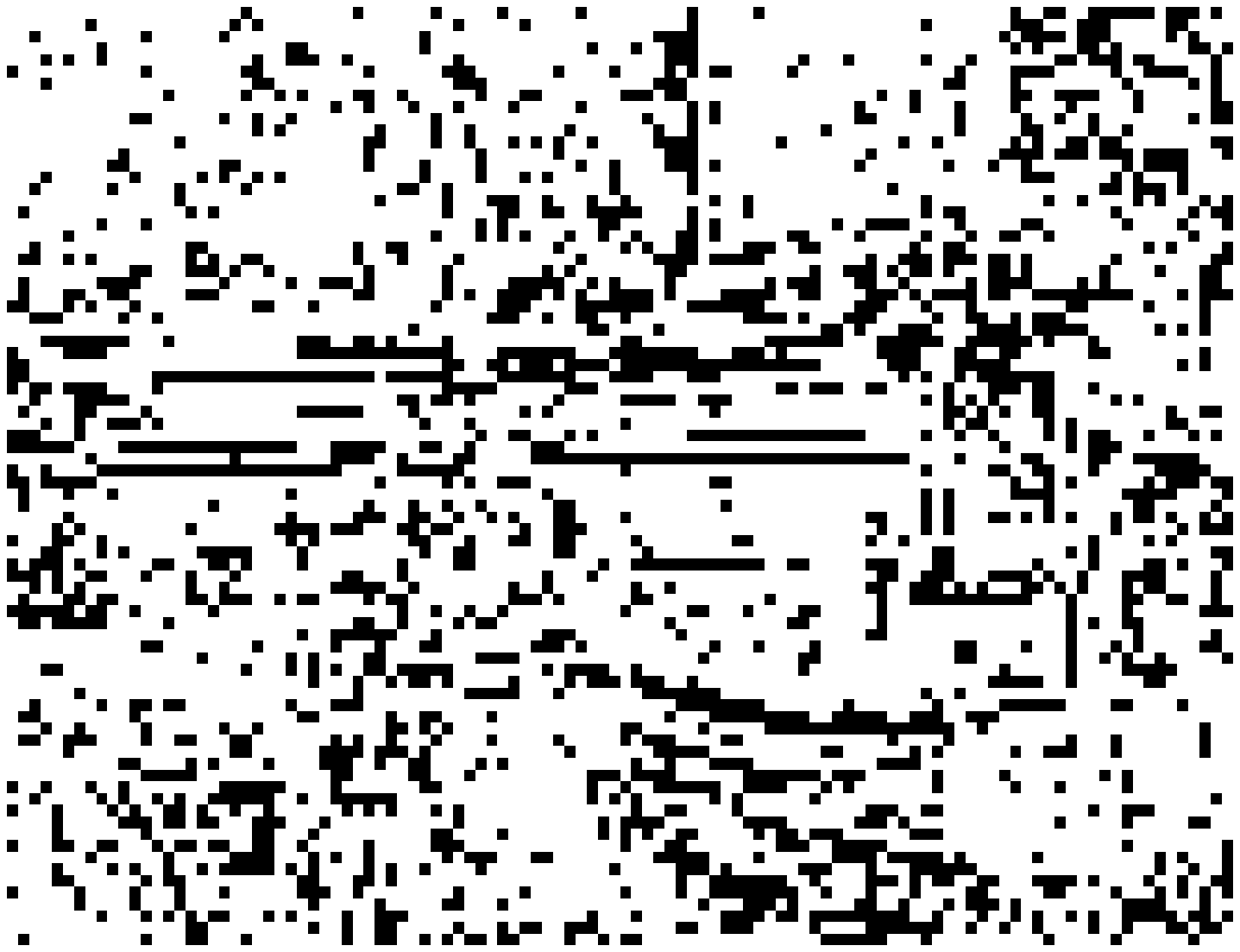,   width=2.7cm,height=2.5cm}}
  \subfigure[]{\epsfig{file=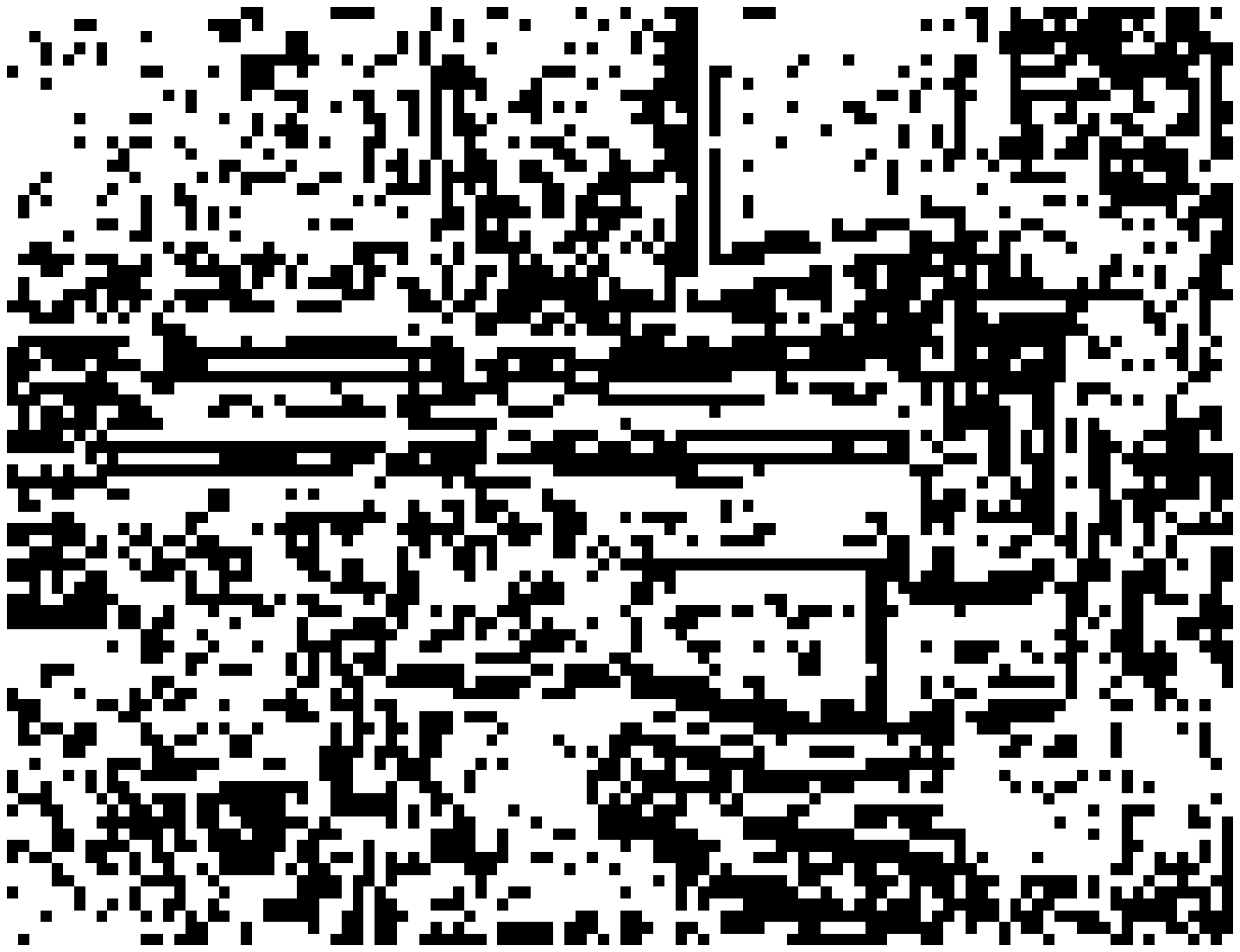,   width=2.7cm,height=2.5cm}}
  \subfigure[]{\epsfig{file=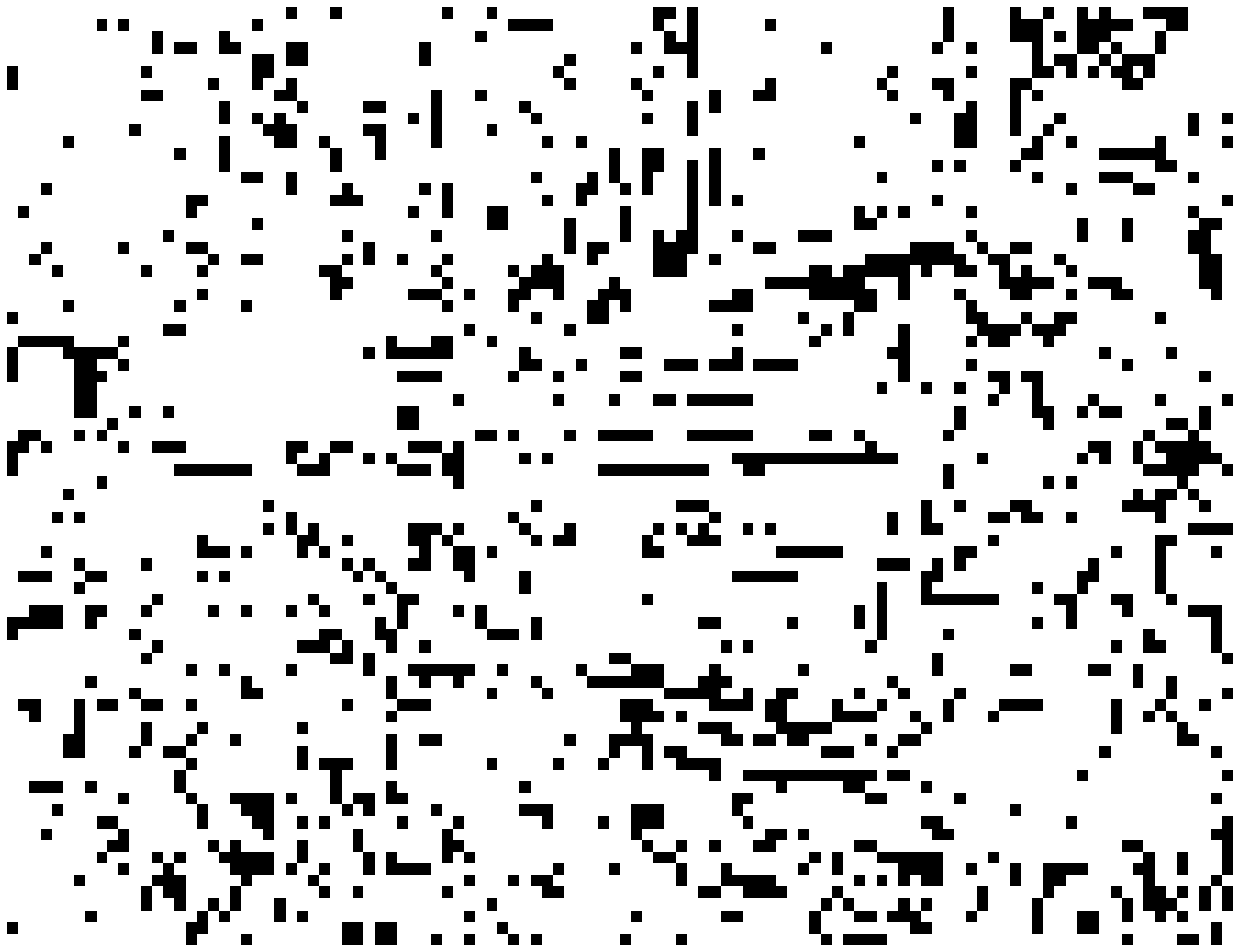, width=2.7cm,height=2.5cm}}
  \subfigure[]{\epsfig{file=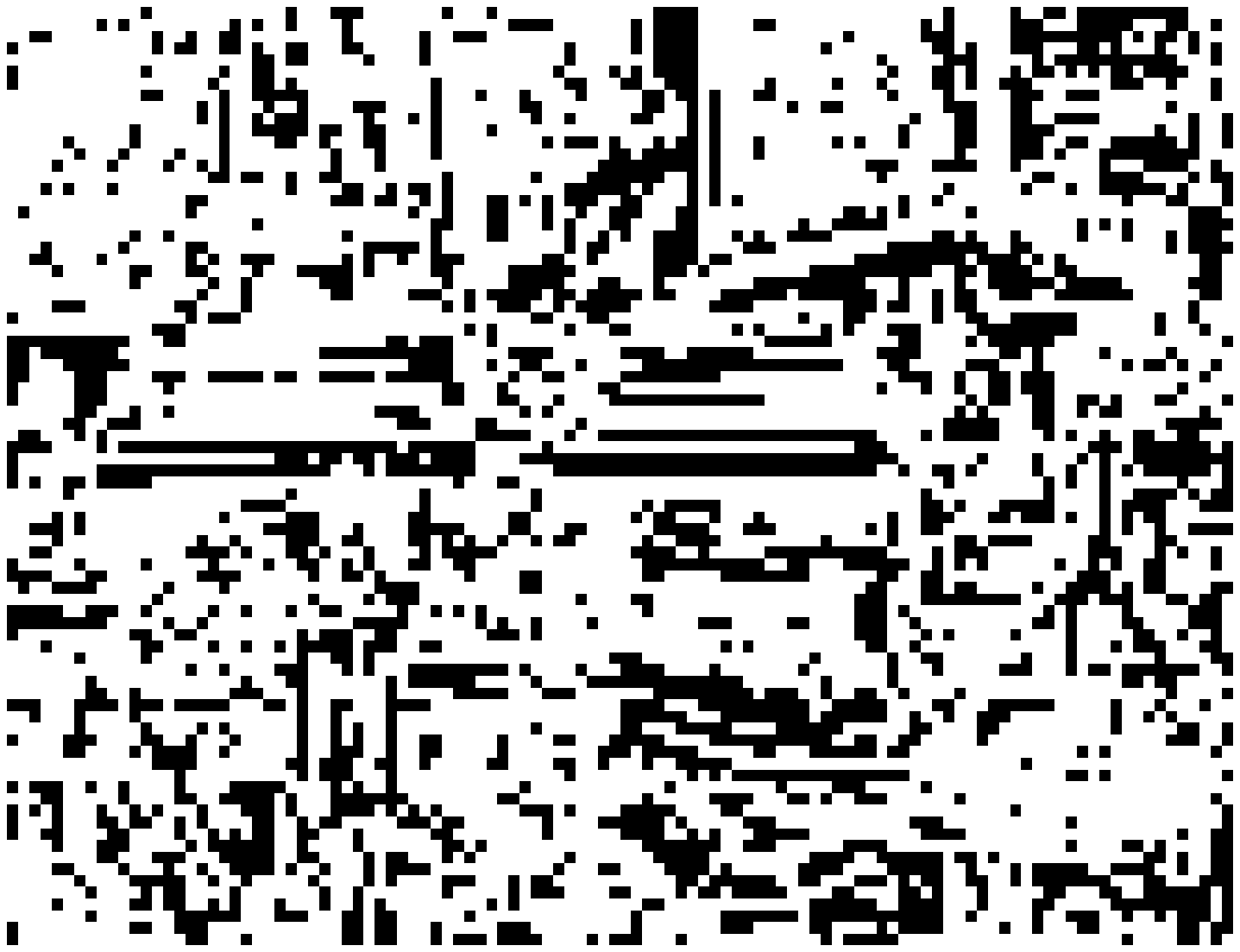,width=2.7cm,height=2.5cm}}
  \subfigure[]{\epsfig{file=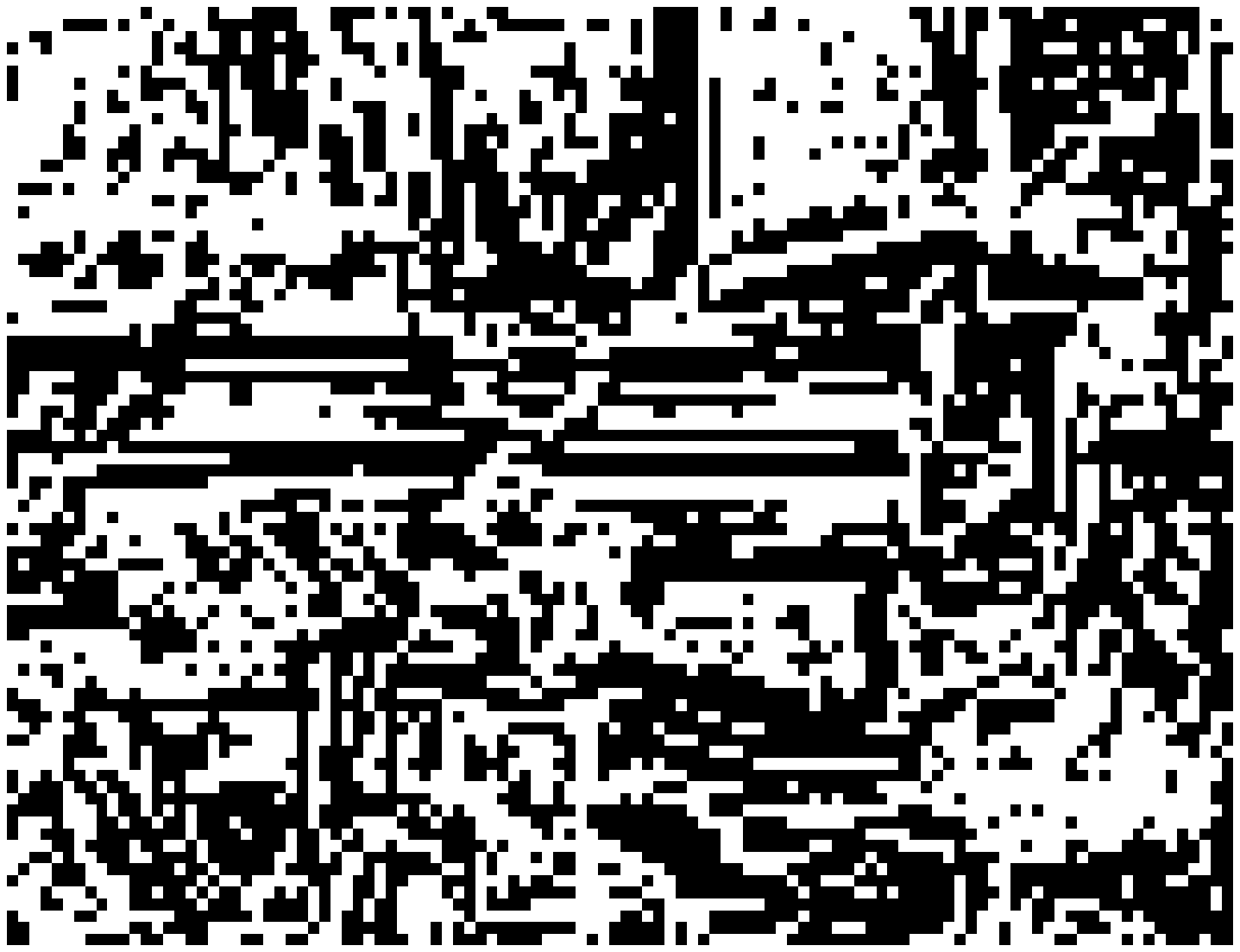,width=2.7cm,height=2.5cm}}
  \caption
        {Reliable points $\mathrm{RP}_\epsilon$
         for invariant $\Theta_{m12\gamma}$, in black, image {\tt WoBA},
	 without and with prefiltering.
        \ (a)~$\sigma_{pre}$=0,  \ $\epsilon$= 5.0;
        \ (b)~$\sigma_{pre}$=0,  \ $\epsilon$=10.0;
        \ (c)~$\sigma_{pre}$=0,  \ $\epsilon$=20.0;
        \ (d)~$\sigma_{pre}$=1.0,\ $\epsilon$= 5.0;
        \ (e)~$\sigma_{pre}$=1.0,\ $\epsilon$=10.0;
        \ (f)~$\sigma_{pre}$=1.0,\ $\epsilon$=20.0.
        }
  \label{fig:ReliaPts}
 \end{center}
\end{figure}

In addition to computing the absolute error, we can also compute the
relative error, in percent, as
\begin{equation}
 \label{eq:RelErr}
  \delta_{CGC}(i,j)=100\ \Delta_{CGC}(i,j)\,/\,\Theta_{0GC}(i,j)
\end{equation}
Then we can define the set $\mathrm{RP}_\epsilon$ of {\em reliable points},
relative to some error threshold~$\epsilon$, as
\begin{equation}
 \label{eq:ReliaPts}
  \mathrm{RP}_\epsilon = \{(i,j)\,|\,\delta(i,j) \leq \epsilon \}
\end{equation}
and $\mathrm{PRP}_\epsilon$, the percentage of reliable points, as
\begin{equation}
 \label{eq:PercReliaPts}
  \mathrm{PRP}_\epsilon = 100\ |\mathrm{RP}_\epsilon|\,/\,n
\end{equation}
where $n$ is the number of valid, i.e.~non-boundary, pixels in the image.
Fig.~\ref{fig:ReliaPts} shows, in the first row, the reliable points
for three different values of the threshold $\epsilon$.
The second row shows the sets of reliable points for the same thresholds
if we gently prefilter
the 0GC and CGC images.
The corresponding data for the ten test images from fig.~\ref{fig:ImaDB}
is summarized in
table~\ref{tab:ReliaPerc}.
\begin{table}[ht]
  \small{
  \begin{center}
  \begin{tabular} {c||c|c|c||c|c|c}
    %image & $\epsilon=5.0$ & $\epsilon=10.0$ & $\epsilon=20.0$
    %      & $\epsilon=5.0$ & $\epsilon=10.0$ & $\epsilon=20.0$
    image & 5.0 & 10.0 & 20.0 & 5.0 & 10.0 & 20.0 \\
    \hline
     {\tt Build}   & 13.3 & 24.9 & 43.8 & 16.0 & 29.5 & 49.3 \\
     {\tt WoBA}    & 15.6 & 29.0 & 48.2 & 19.0 & 35.7 & 58.9 \\
     {\tt WoBB}    & 16.5 & 28.7 & 47.1 & 21.4 & 37.7 & 58.1 \\
     {\tt WoBC}    & 18.5 & 33.6 & 53.5 & 24.0 & 41.4 & 65.3 \\
     {\tt WoBD}    & 13.0 & 23.9 & 41.9 & 16.7 & 32.6 & 55.6 \\
     {\tt Cycl}    & 15.4 & 28.3 & 45.9 & 22.6 & 38.8 & 57.6 \\
     {\tt Sand}    & 14.5 & 27.2 & 44.7 & 22.0 & 38.5 & 57.6 \\
     {\tt ToolA}   &  5.6 & 10.7 & 20.1 &  7.4 & 14.7 & 27.1 \\
     {\tt ToolB}   &  6.1 & 12.0 & 22.7 &  8.3 & 15.7 & 28.6 \\
     {\tt ToolC}   &  5.6 & 11.1 & 20.8 &  7.9 & 15.1 & 28.3 \\
    \hline
     median	   & 13.9 & 26.1 & 44.3 & 17.9 & 34.2 & 56.6 \\
     mean	   & 12.4 & 22.9 & 38.9 & 16.5 & 30.0 & 48.6
  \end{tabular}
  \end{center}
  }
  \caption
        {Percentages of reliable points for $\Theta_{m12\gamma}$, CGC images,
         for $\epsilon$=5.0, 10.0, 20.0.
         The three numerical columns on the left show $\mathrm{PRP}_\epsilon$
         without prefiltering, the three right columns with 
	 Gaussian prefiltering, $\sigma_{pre}$=1.0.
        }
 \label{tab:ReliaPerc}
\end{table}

Derivatives are known to be sensitive to noise.
Noise can be reduced by smoothing the original data
before the invariants are computed.
On the other hand, derivatives should be computed as locally as possible.
With these conflicting goals to be considered,
we experiment with gentle prefiltering,
using a Gaussian filter of size $\sigma_{pre}$=1.0.
The size of the Gaussian to compute the invariant $\Theta_{m12\gamma}$
is set to $\sigma_{der}$=1.0.
Note that $\sigma_{pre}$ and $\sigma_{der}$ can {\em not} be combined
into just one Gaussian because of the non-linearity of the invariant.

With respect to the set of reliable points,
we observe that after prefiltering, roughly half the points, on average,
have a relative error of less than 20\%.
Gentle prefiltering consistently reduces both absolute and relative
errors, but strong prefiltering does not.

%--------------------------------------------------------------------------
\subsection{Template Matching}
%--------------------------------------------------------------------------

Template matching is a frequently employed technique in computer vision.
Here, we will examine how gamma correction affects the spatial accuracy of
template matching, and whether that accuracy can be improved by
using the invariant $\Theta_{m12\gamma}$.
\begin{figure}[htbp]
 \begin{center}
  \epsfig{file=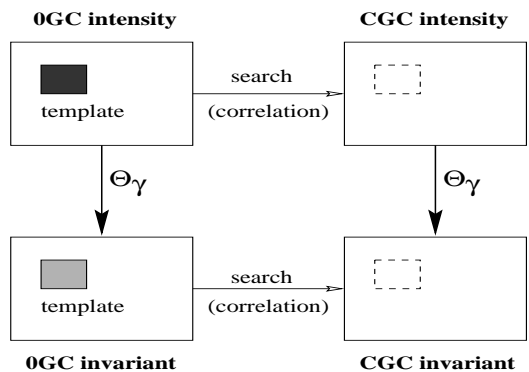,width=6.9cm,height=4.8cm}
  \caption
        {The template location problem:
         A query template is cut out from the 0GC intensity image and
         correlated with the corresponding CGC intensity image.
         We test if the correlation maximum occurs at exactly the
         same location as in the 0GC intensity image.
         The same process is repeated with the invariant representations
         of the 0GC and CGC images.
        }
  \label{fig:TemplLoca}
 \end{center}
\end{figure}
An overview of the testbed scenario is given in fig.~\ref{fig:TemplLoca}.
A~small template of size $6 \times 8$, representing the search pattern,
is taken from a 0GC intensity image, i.e.~without gamma correction.
This query template is then correlated
with the corresponding CGC intensity image,
i.e.~the same scene but with gamma correction switched on.
If the correlation maximum occurs at exactly the location
where the 0GC query template has been cut out, we call this a
{\em correct maximum correlation position}, or CMCP.

The correlation function $s(x,y)$
employed here is based on a
normalized mean squared difference $c(x,y)$~\cite{fua93}:
\[
 \begin{aligned}
  s &= \max (0, 1-c) \\
  c &= \frac
        {\sum_{i,j} ((I(x+i,y+j)-\overline{I}) -
                (T(i,j)-\overline{T}))^2}
        {\sqrt{\sum_{i,j} (I(x+i,y+j)-\overline{I})^2 \
               \sum_{i,j} (T(i,j)-\overline{T})^2}}
 \end{aligned}
\]
where $I$ is an image, $T$ is a template positioned at $(x,y)$,
$\overline{I}(x,y)$ is the mean of the subimage of $I$ at $(x,y)$
of the same size as $T$,
$\overline{T}$ is the mean of the template,
and $0 \leq s \leq 1$.
The template location problem then is to perform this correlation
for the whole image and to determine 
whether the position of the correlation maximum occurs precisely at $(x,y)$.

\begin{figure}[htbp]
 \begin{center}
  \epsfig{file=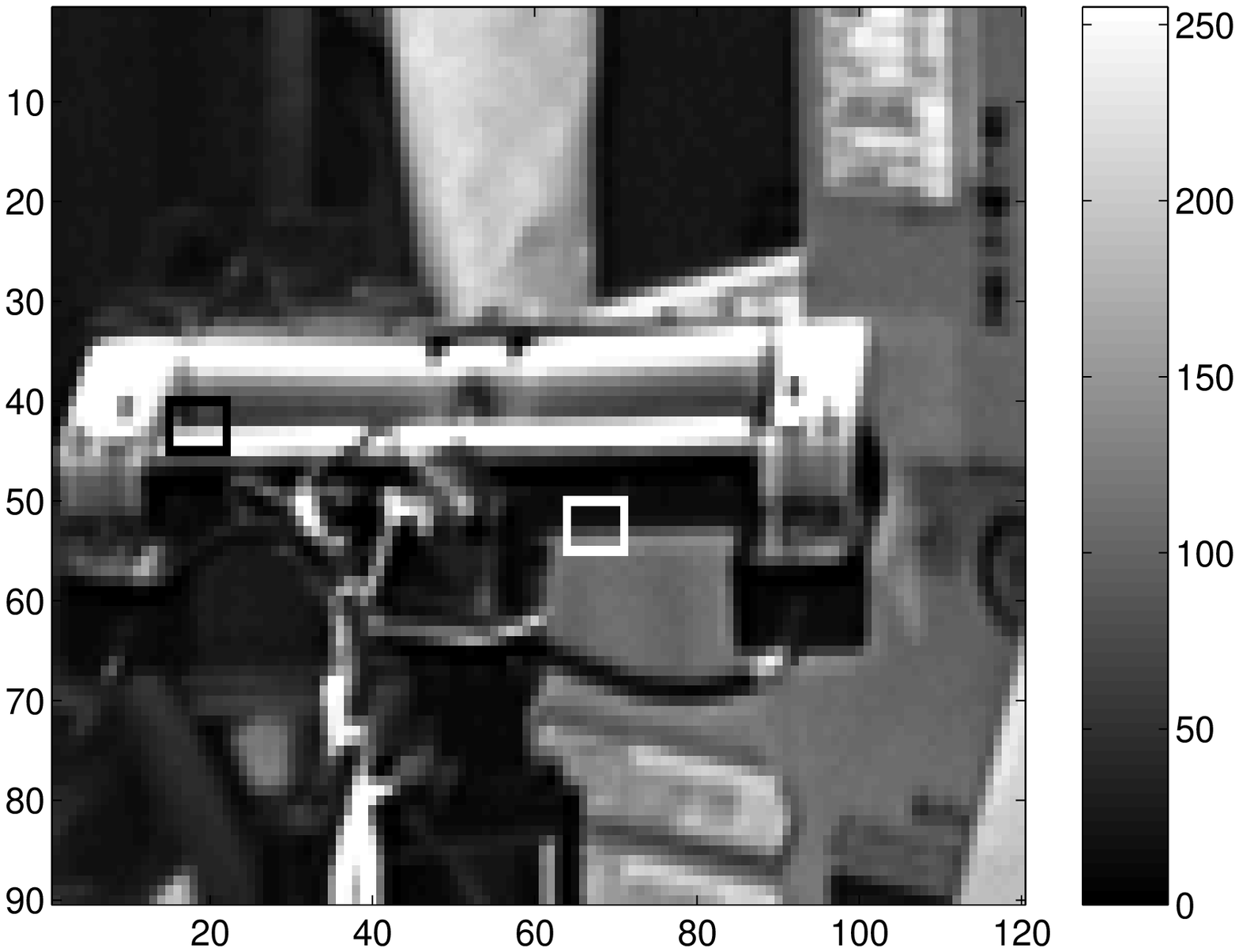,     width=4.1cm}
  \epsfig{file=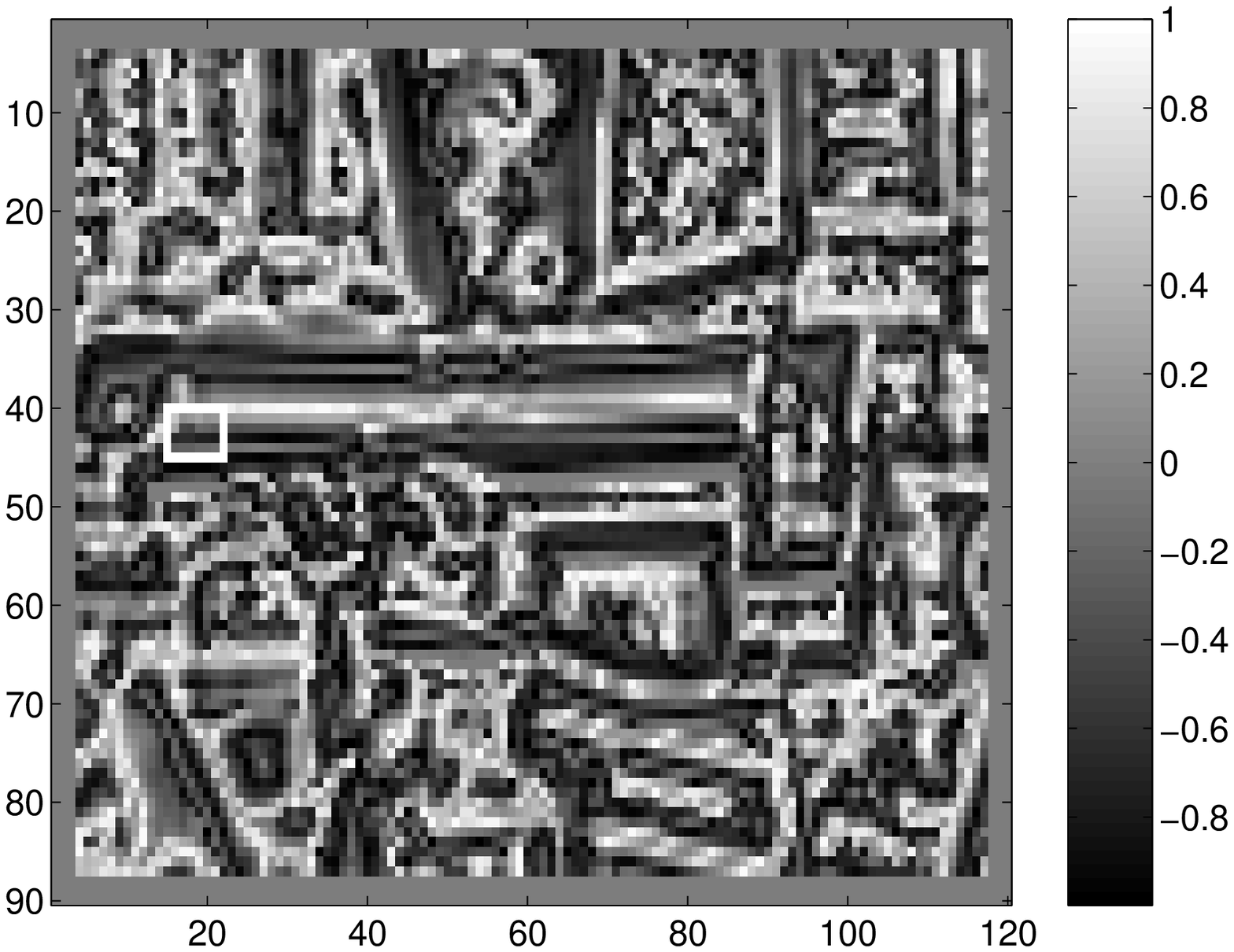,width=4.1cm}
  \caption
	{Matched templates, image {\tt WoBA}:
	 \ (left) intensity data;
	 \ (right) invariant representation.
	 Black box=query template, white box=matched template.
        }
  \label{fig:MatchTempl}
 \end{center}
\end{figure}

\begin{figure}[htbp]
 \begin{center}
  \epsfig{file=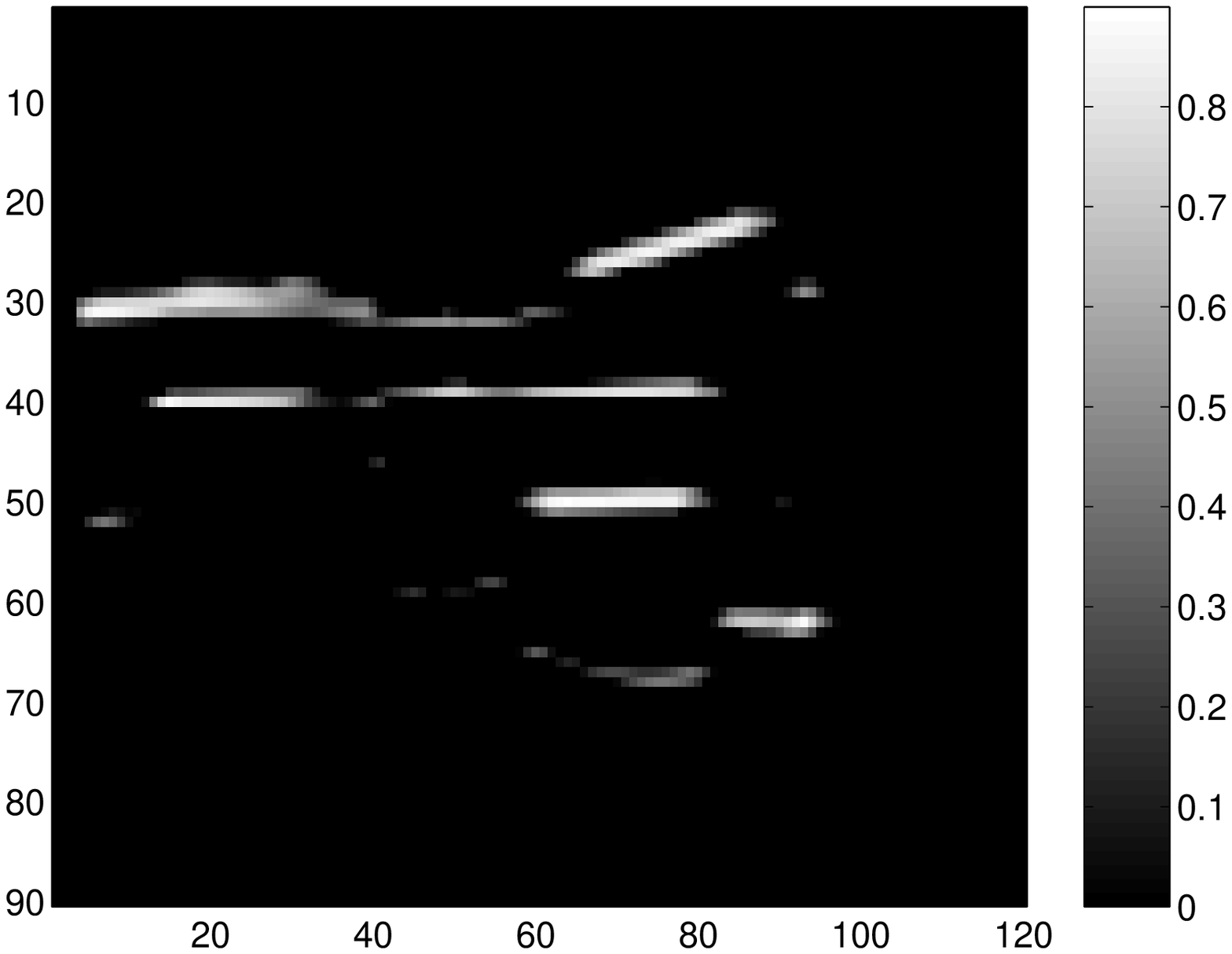,        width=4.1cm}
  \epsfig{file=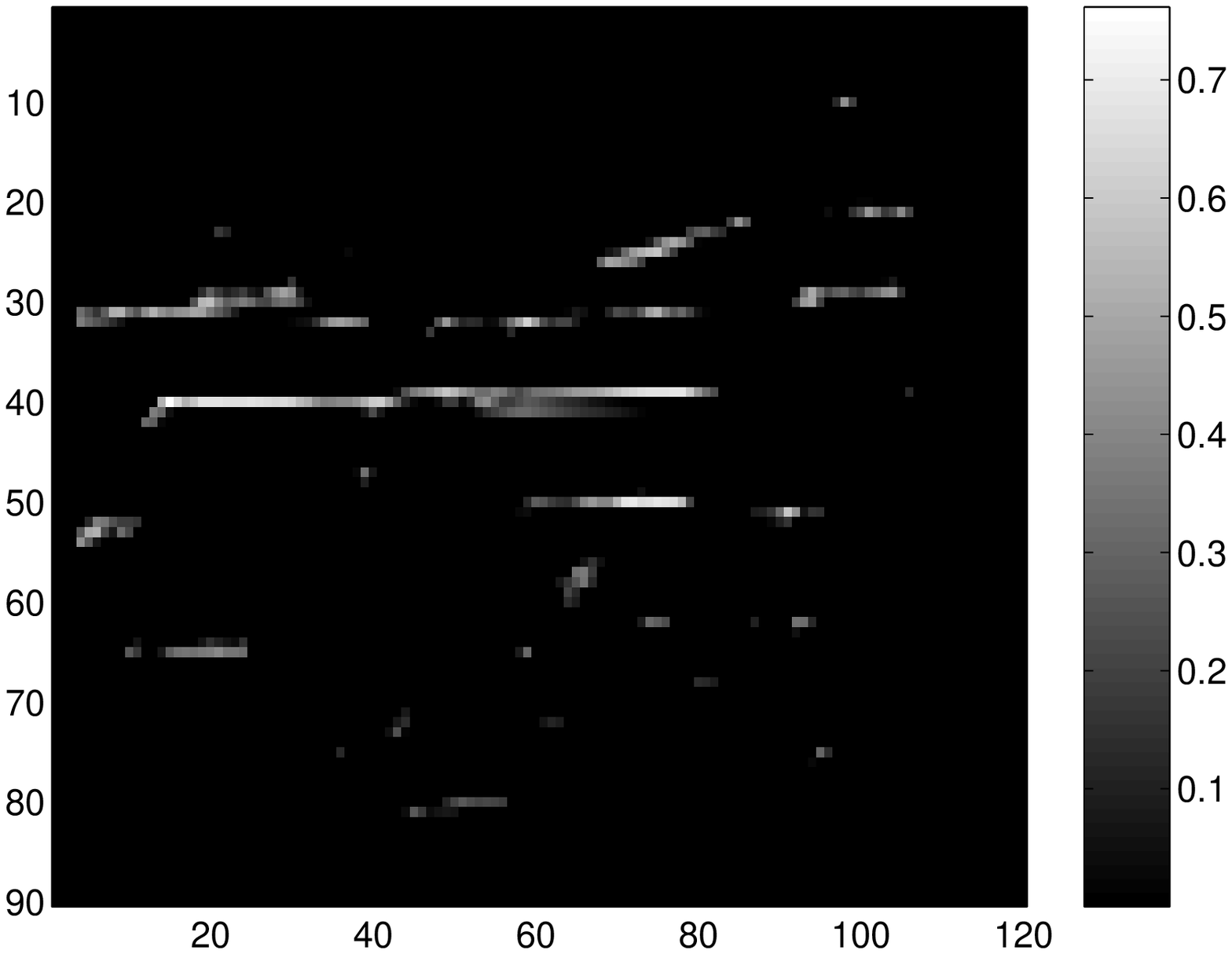,width=4.1cm}
  \caption
	{Correlation matrices, image {\tt WoBA}.
	 \ (left) intensity data;
	 \ (right) invariant representation.
        }
  \label{fig:CorrelExmpl}
 \end{center}
\end{figure}
Fig.~\ref{fig:MatchTempl} demonstrates the template location problem,
on the left for an intensity image, and on the right for its 
invariant representation.
The black box marks the position of the original template at (40,15),
and the white box marks the position of the matched template,
which is incorrectly located at (50,64) in the intensity image.
On the right, the matched template (white) has overwritten
the original template (black) at the same, correctly identified position.
Fig.~\ref{fig:CorrelExmpl} visualizes the correlation function
over the whole image.
The white areas are regions of high correlation.

\begin{figure}[htbp]
 \begin{center}
  \subfigure[]{\epsfig{file=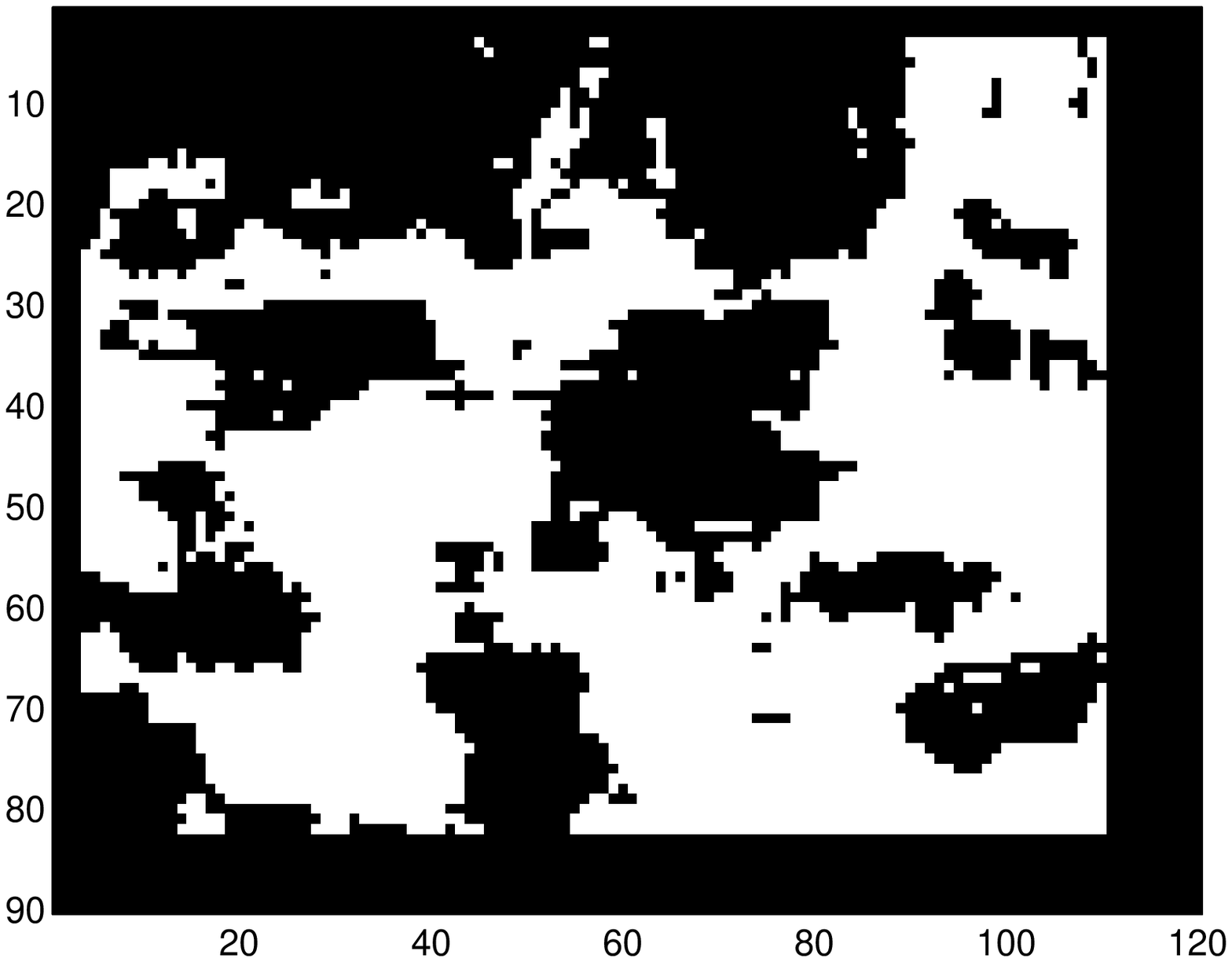,       width=4.1cm}}
  \subfigure[]{\epsfig{file=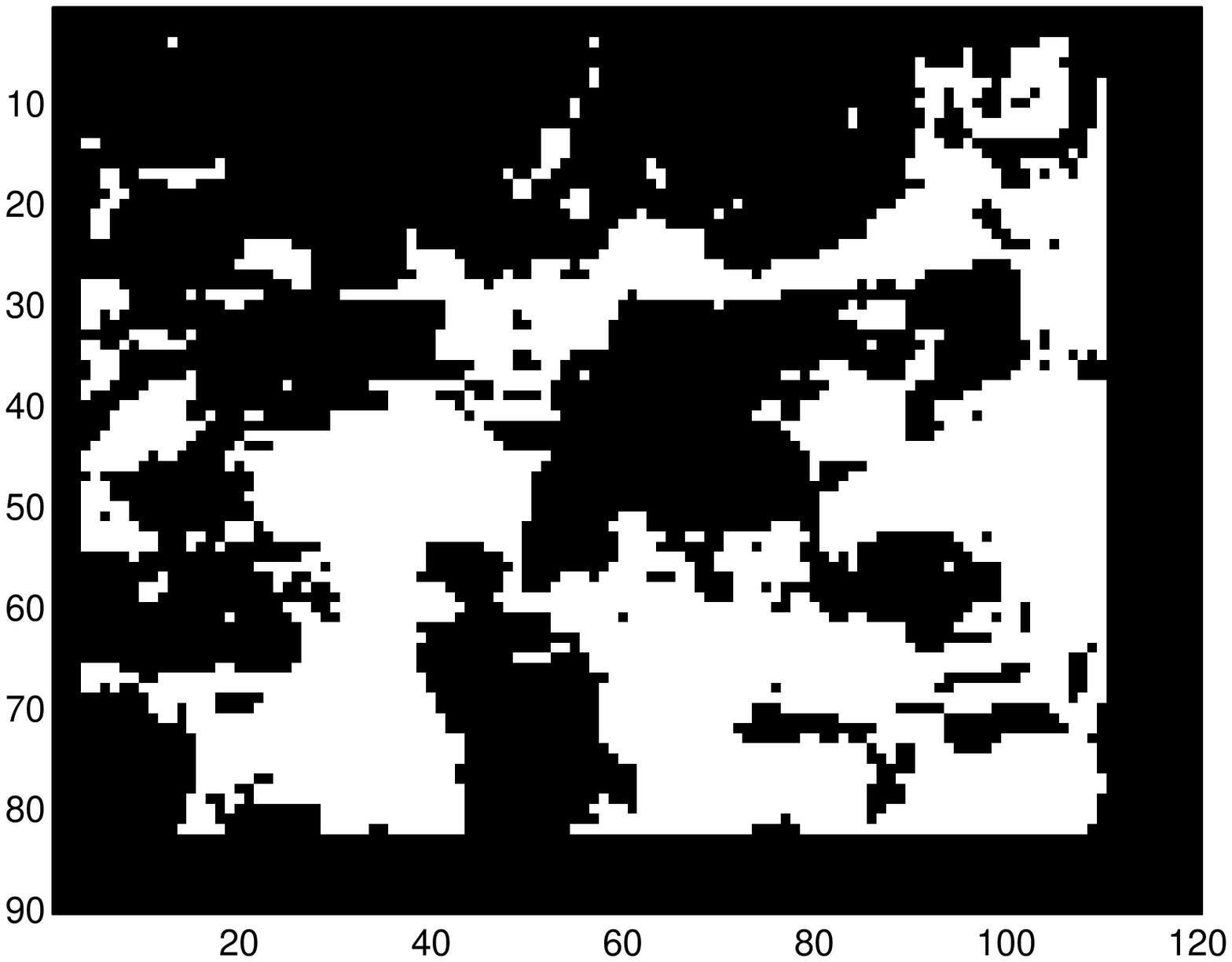,      width=4.1cm}}
  \subfigure[]{\epsfig{file=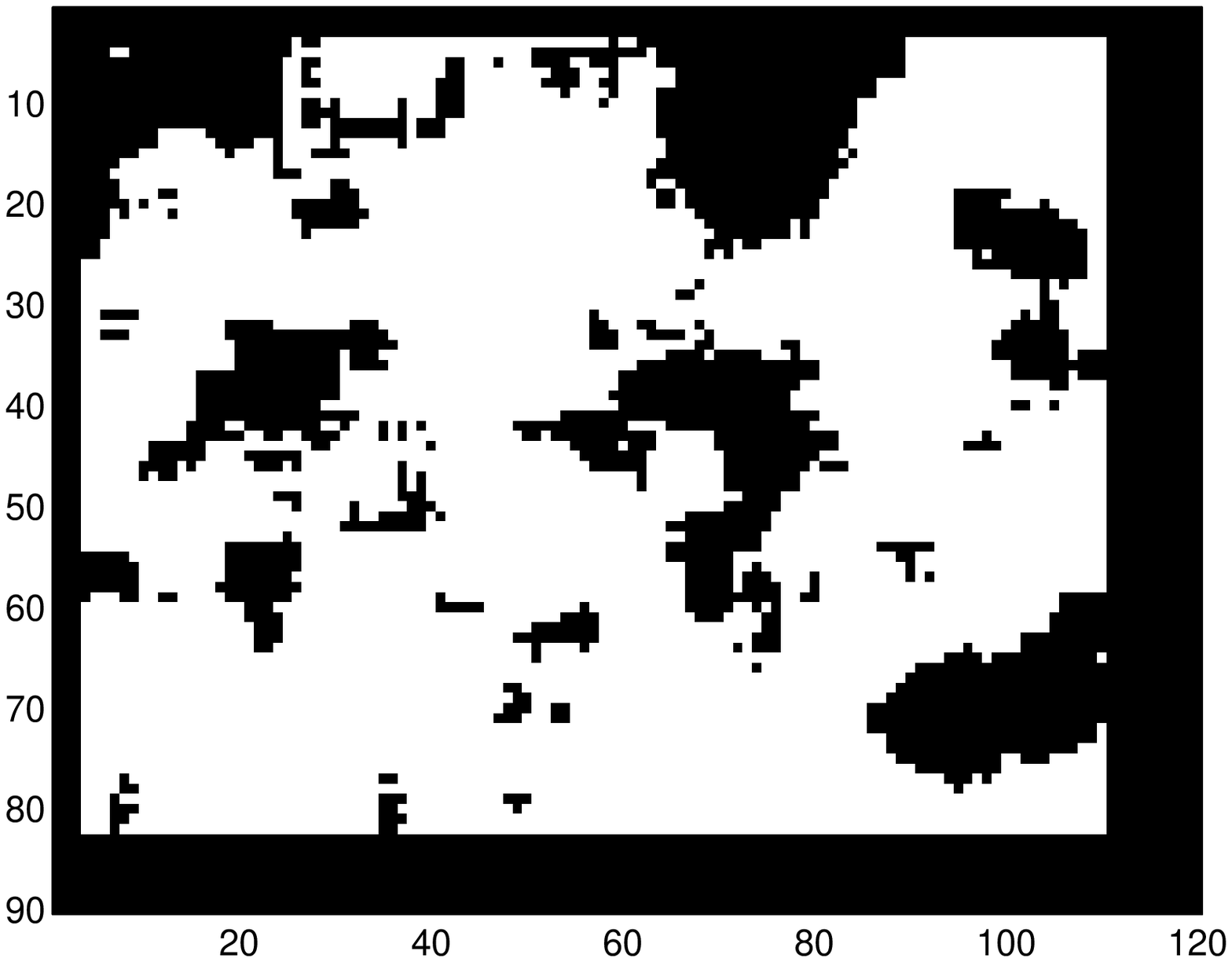, width=4.1cm}}
  \subfigure[]{\epsfig{file=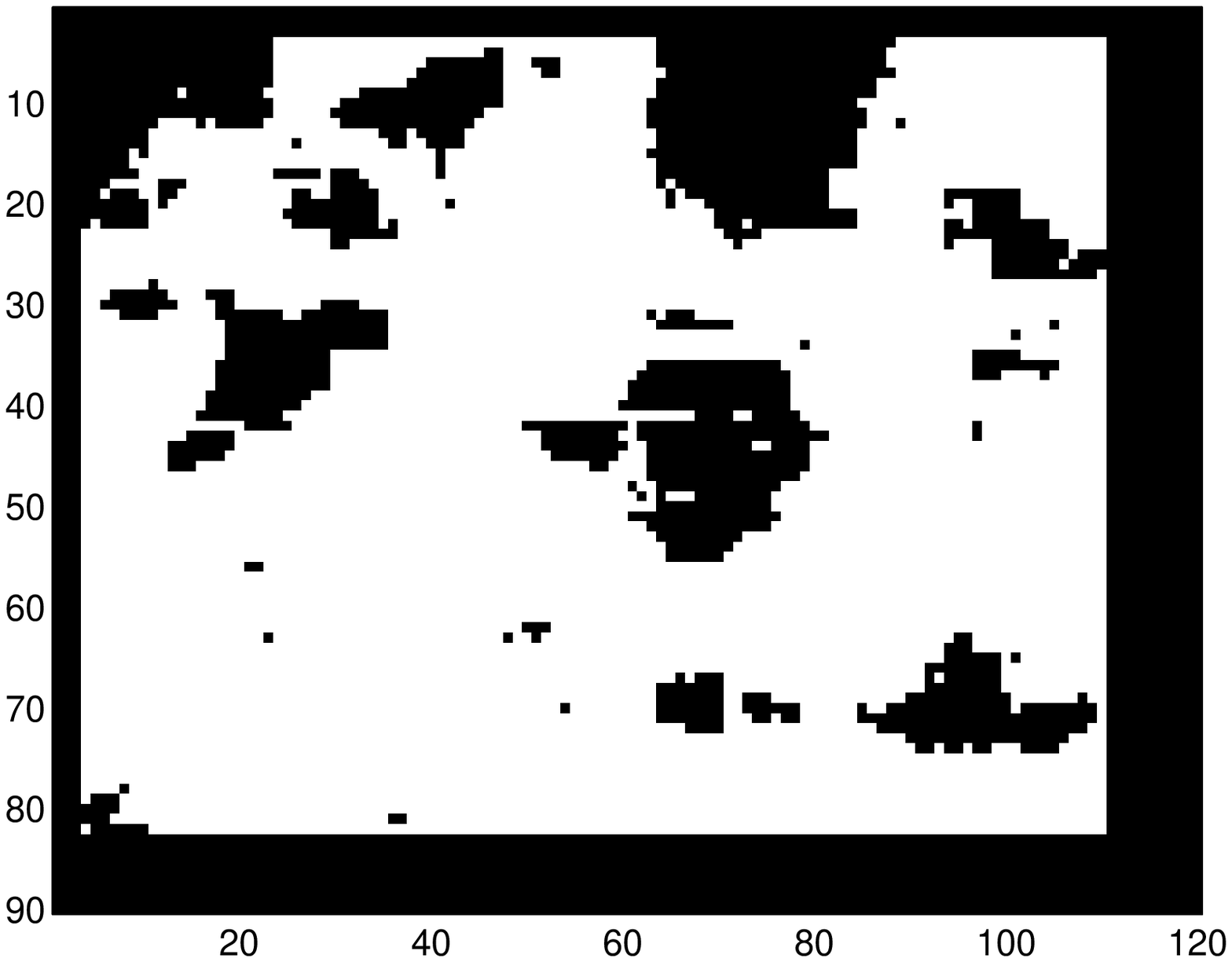,width=4.1cm}}
  \caption
        {Binary correlation accuracy matrices,
         %template size $6 \times 8$,
         white pixels=$CMCP_{6 \times 8}$, image {\tt WoBA}.
        \ (a)~intensity image, $\sigma_{pre}$=0;
        \ (b)~intensity image, $\sigma_{pre}$=1.0;
        \ (c)~invariant representation, $\sigma_{pre}$=0;
        \ (d)~invariant representation, $\sigma_{pre}$=1.0.
        }
  \label{fig:CorrCorrelPts}
 \end{center}
\end{figure}
The example from figs.~\ref{fig:MatchTempl} and~\ref{fig:CorrelExmpl}
deals with only {\em one} arbitrarily selected template.
In order to systematically analyze the template location problem,
we repeat the correlation process for all possible template locations.
Then we can define the {\em correlation accuracy} CA as the
percentage of correctly located templates,
\begin{equation}
 \label{eq:CorrAccura}
  \mathrm{CA}_{tn \times tm}= 100\ |\mathrm{CMCP}_{tn \times tm}|\,/\,n
\end{equation}
where $tn \times tm$ is the size of the template,
$\mathrm{CMCP}_{tn \times tm}$ is the set of
correct maximum correlation positions,
and $n$, again, is the number of valid pixels.
We compute the correlation accuracy both for unfiltered images and for
gently prefiltered images, with $\sigma_{pre}=1.0$.
%Fig.~\ref{fig:CorrCorrelPts} shows the results for our example image.
Fig.~\ref{fig:CorrCorrelPts} shows the binary correlation accuracy
matrices for our example image.
The CMCP set is shown in white, its complement and the boundaries in black.
We observe a higher correlation accuracy for the invariant representation,
which is improved by the prefiltering.

\begin{figure}[htbp]
 \begin{center}
   \epsfig{file=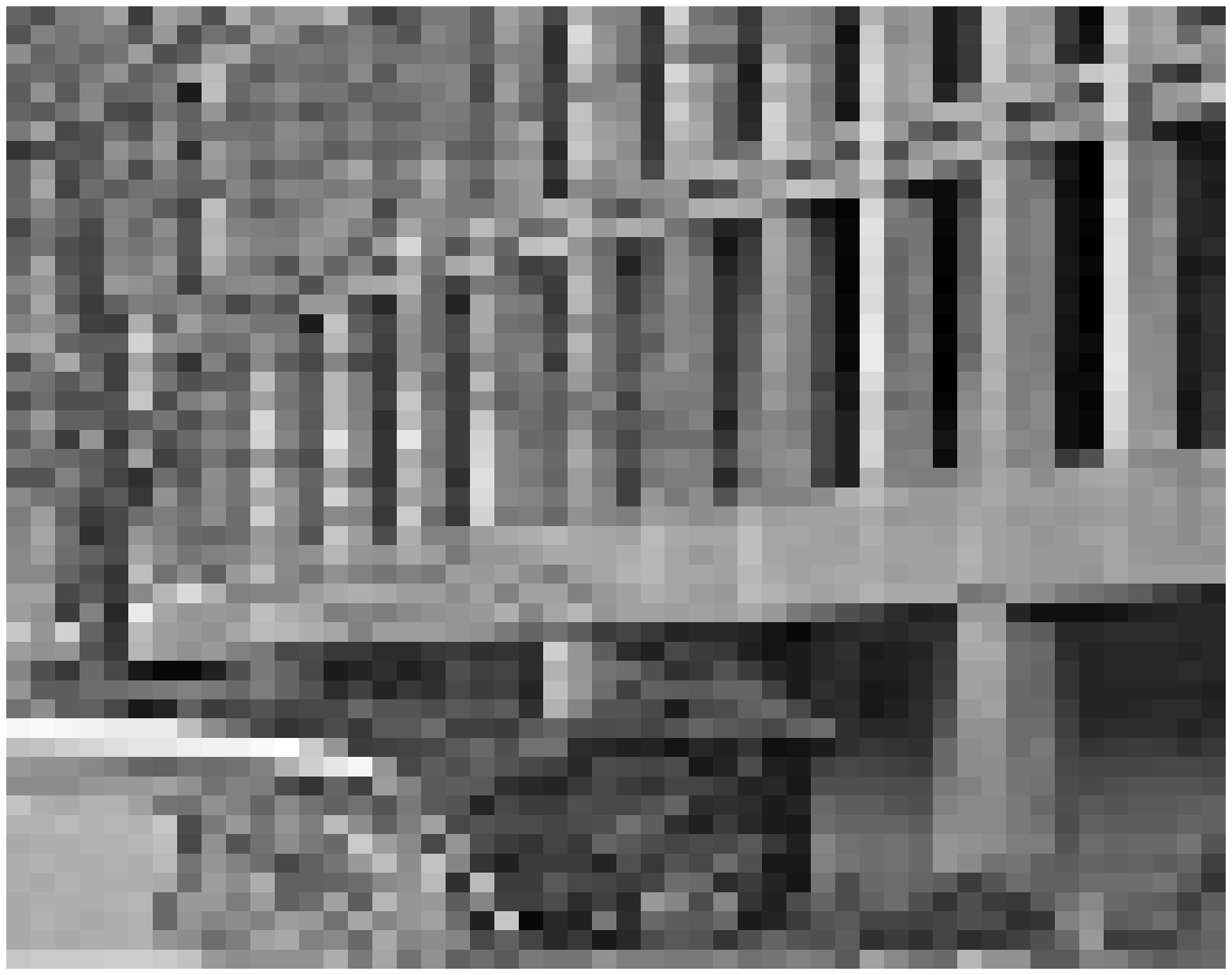,width=1.60cm, height=1.60cm}
   \epsfig{file=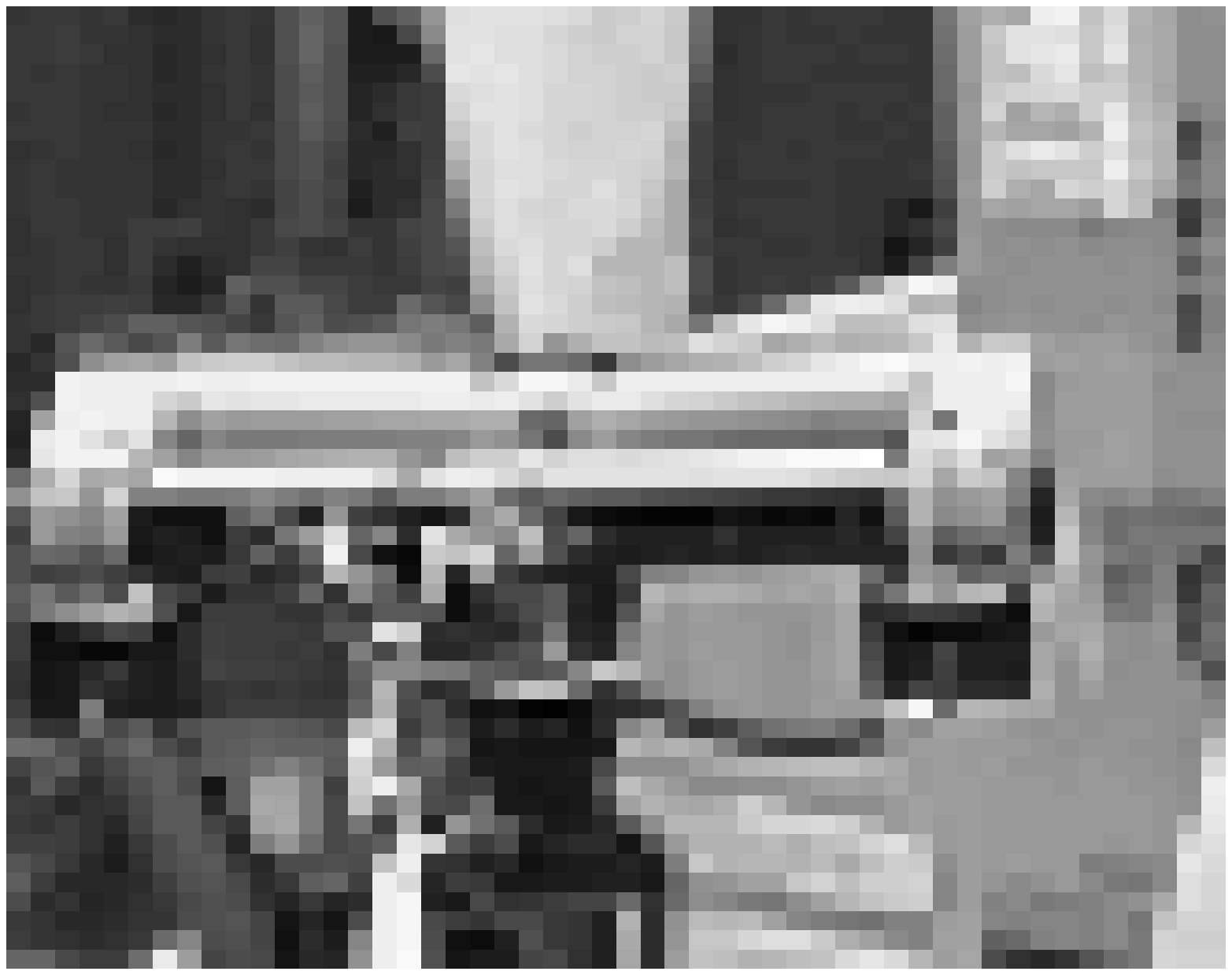,          width=1.60cm, height=1.60cm}
   \epsfig{file=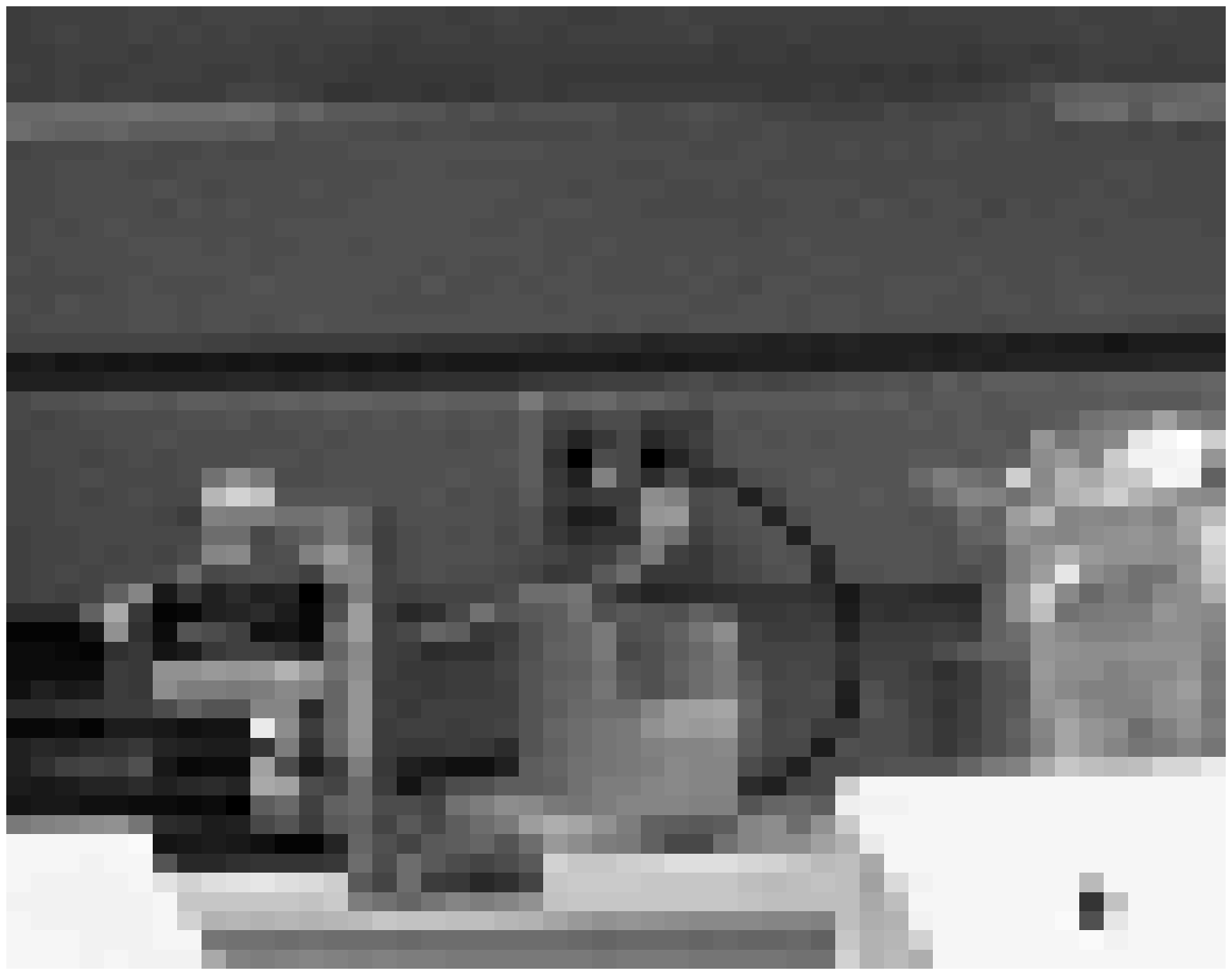,          width=1.60cm, height=1.60cm}
   \epsfig{file=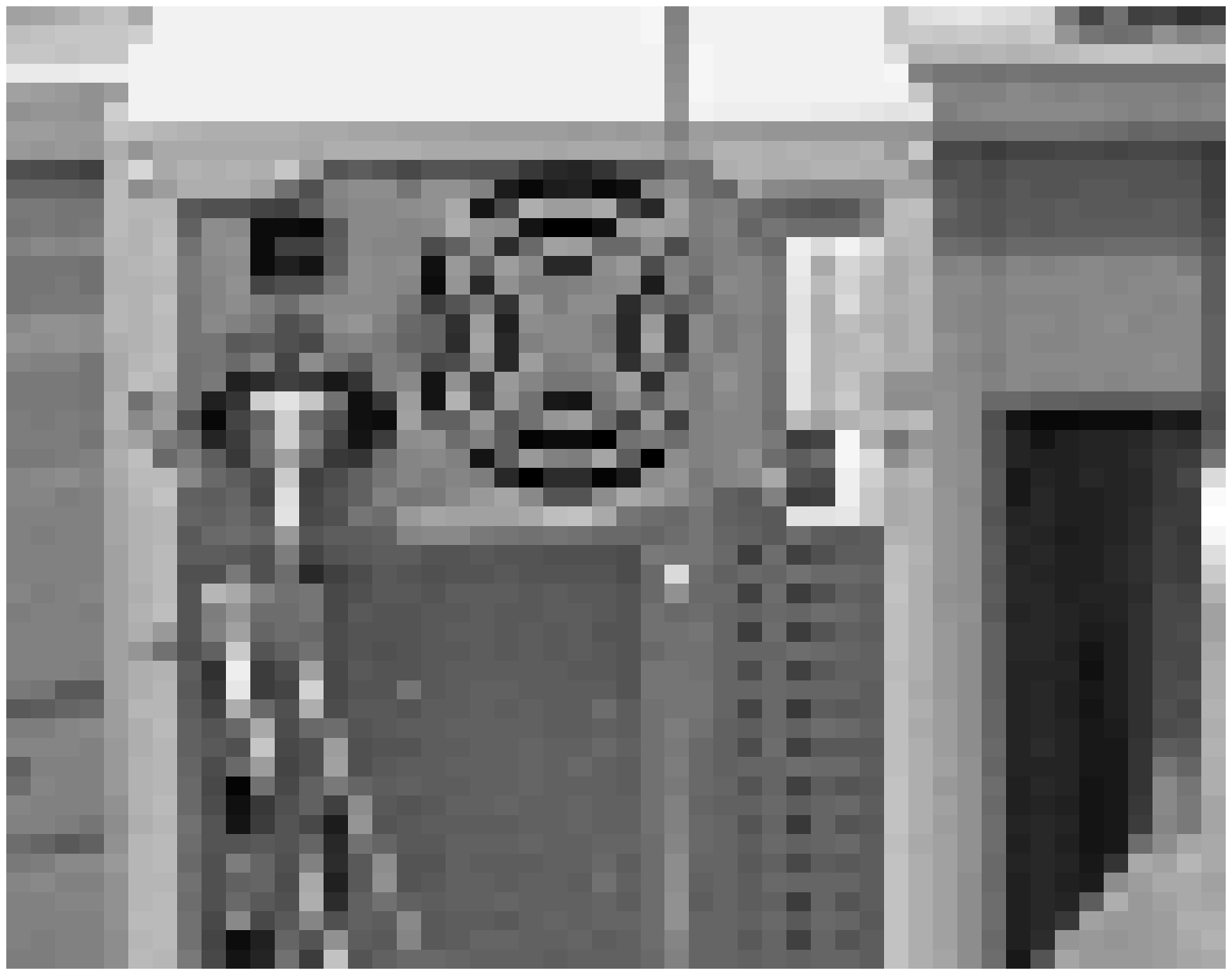,          width=1.60cm, height=1.60cm}
   \epsfig{file=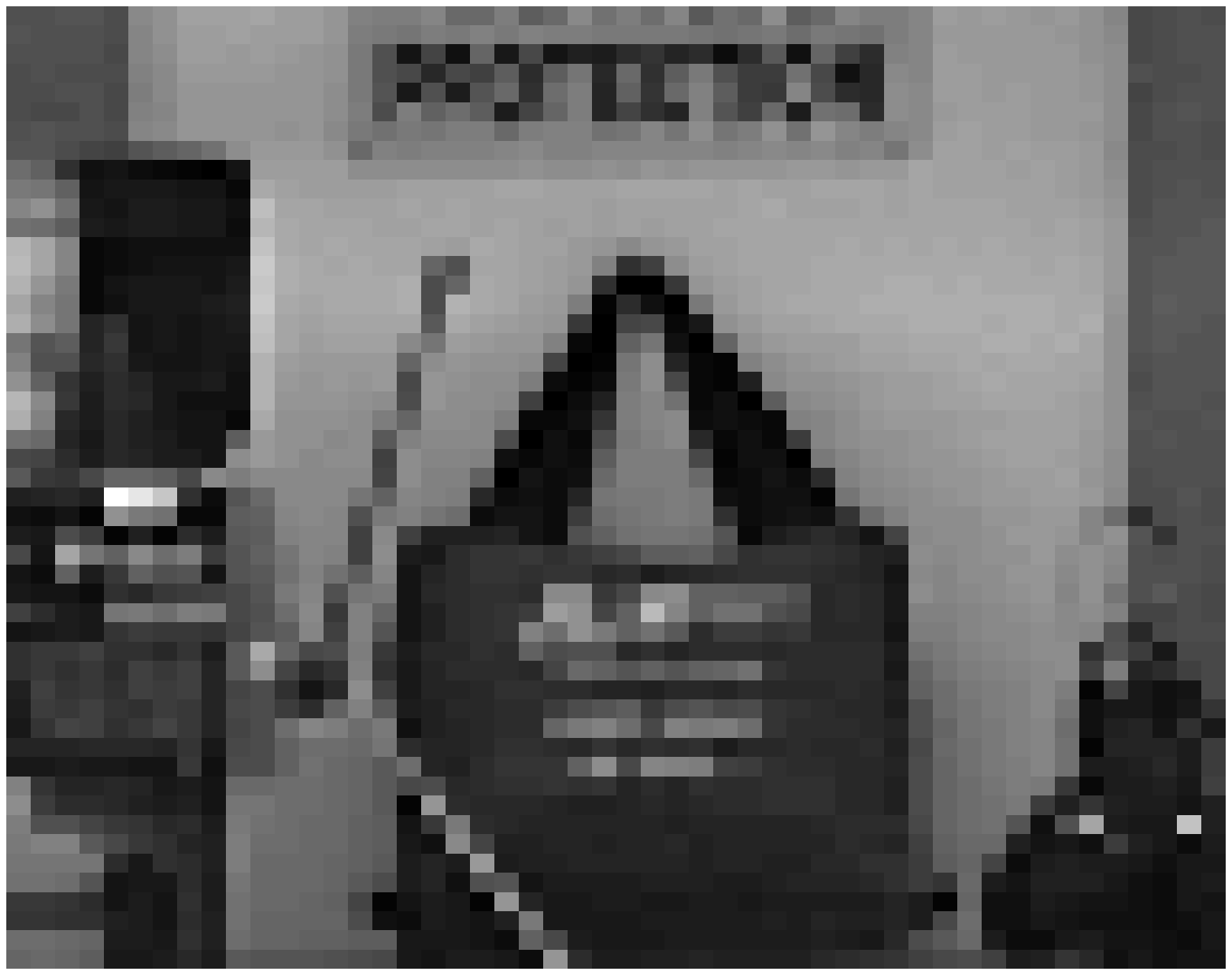,          width=1.60cm, height=1.60cm}
   \epsfig{file=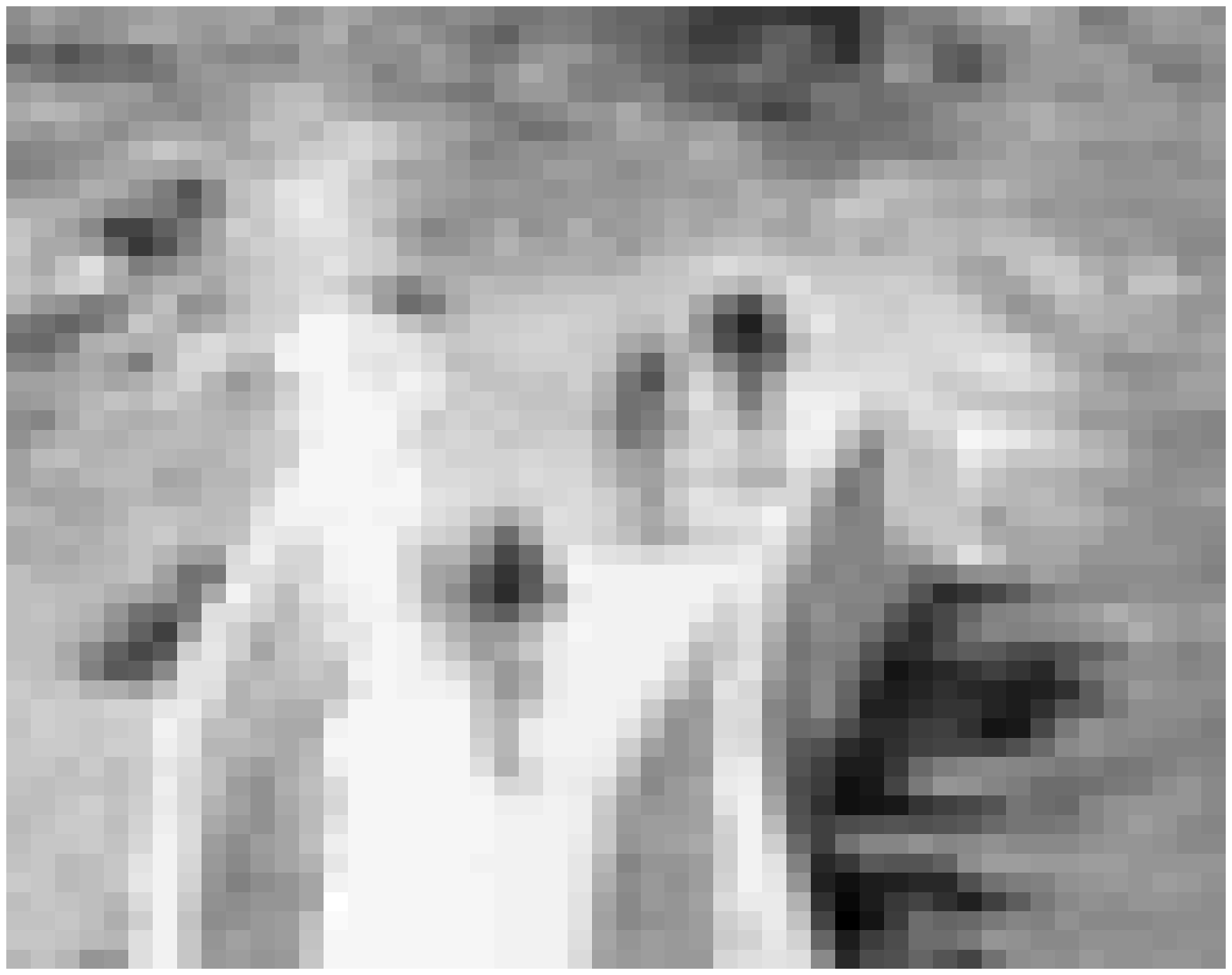,       width=1.60cm, height=1.60cm}
   \epsfig{file=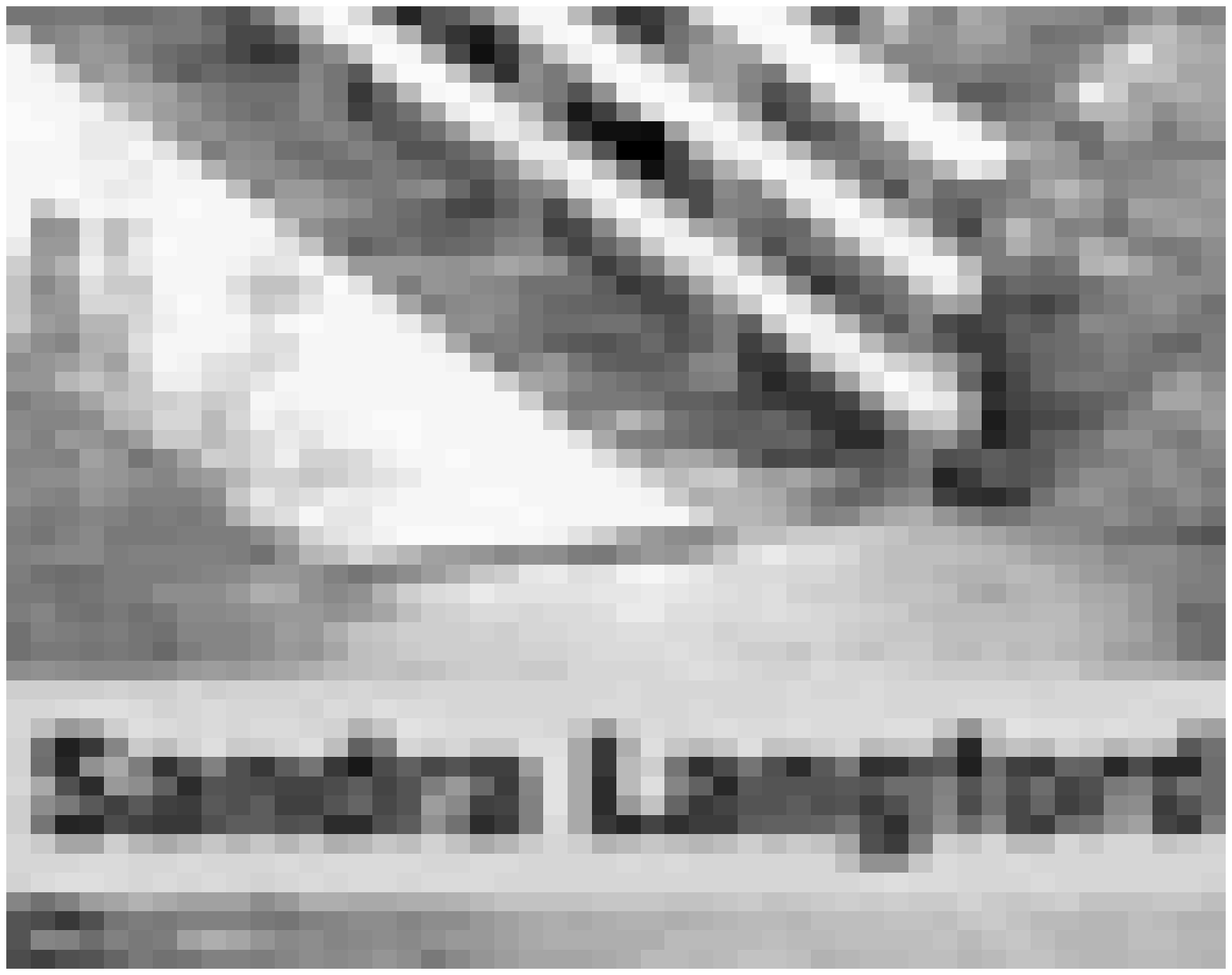,       width=1.60cm, height=1.60cm}
   \epsfig{file=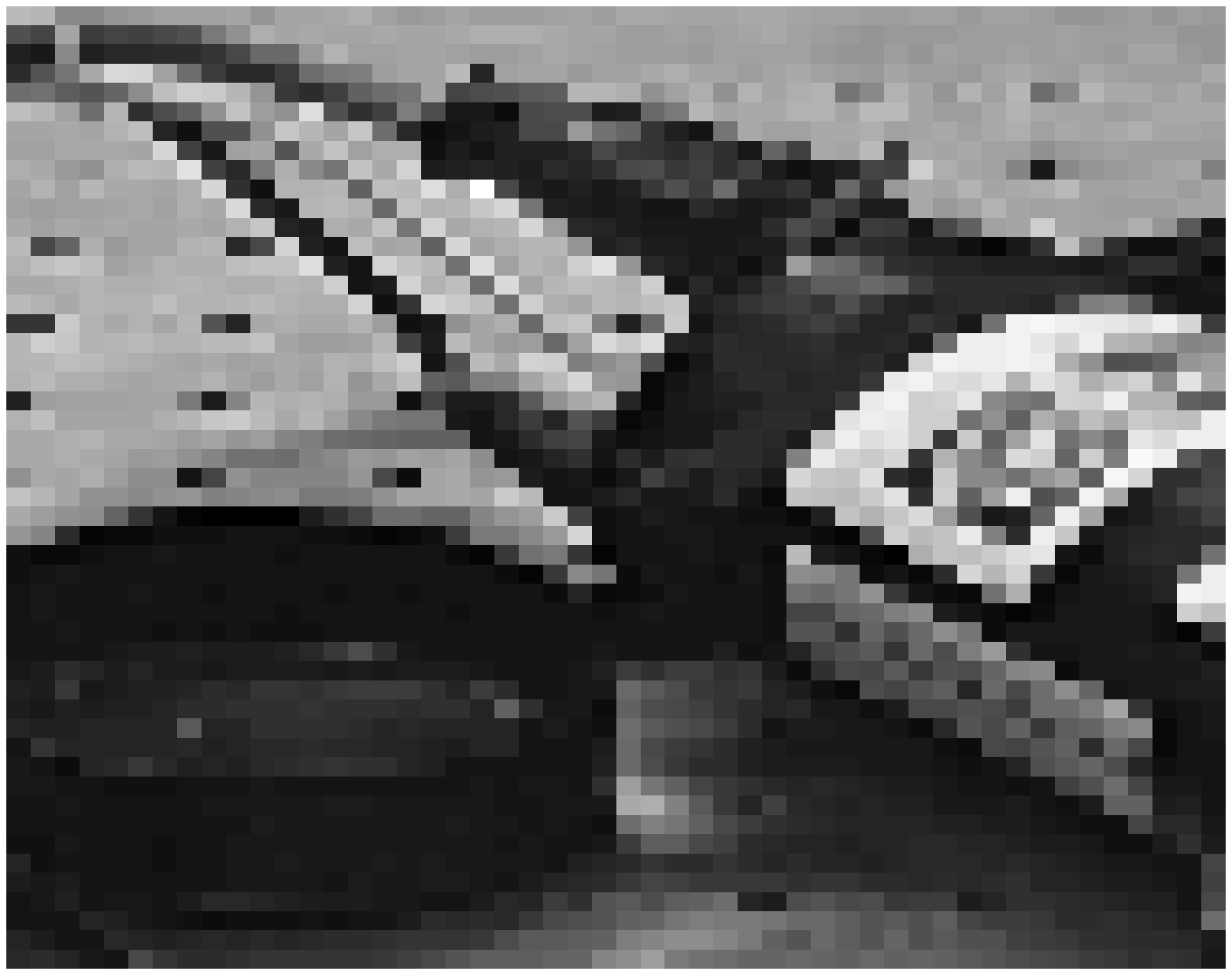,         width=1.60cm, height=1.60cm}
   \epsfig{file=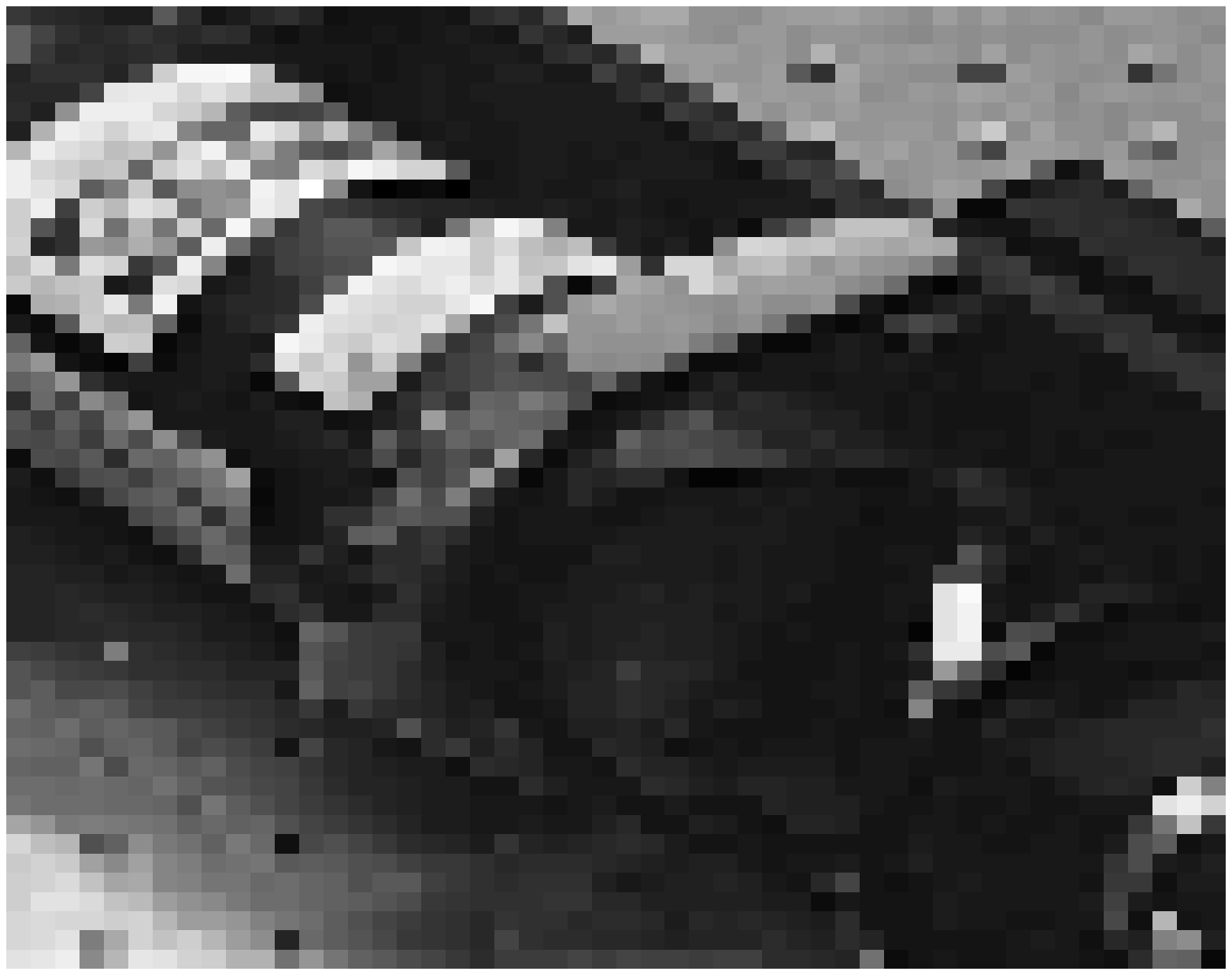,         width=1.60cm, height=1.60cm}
   \epsfig{file=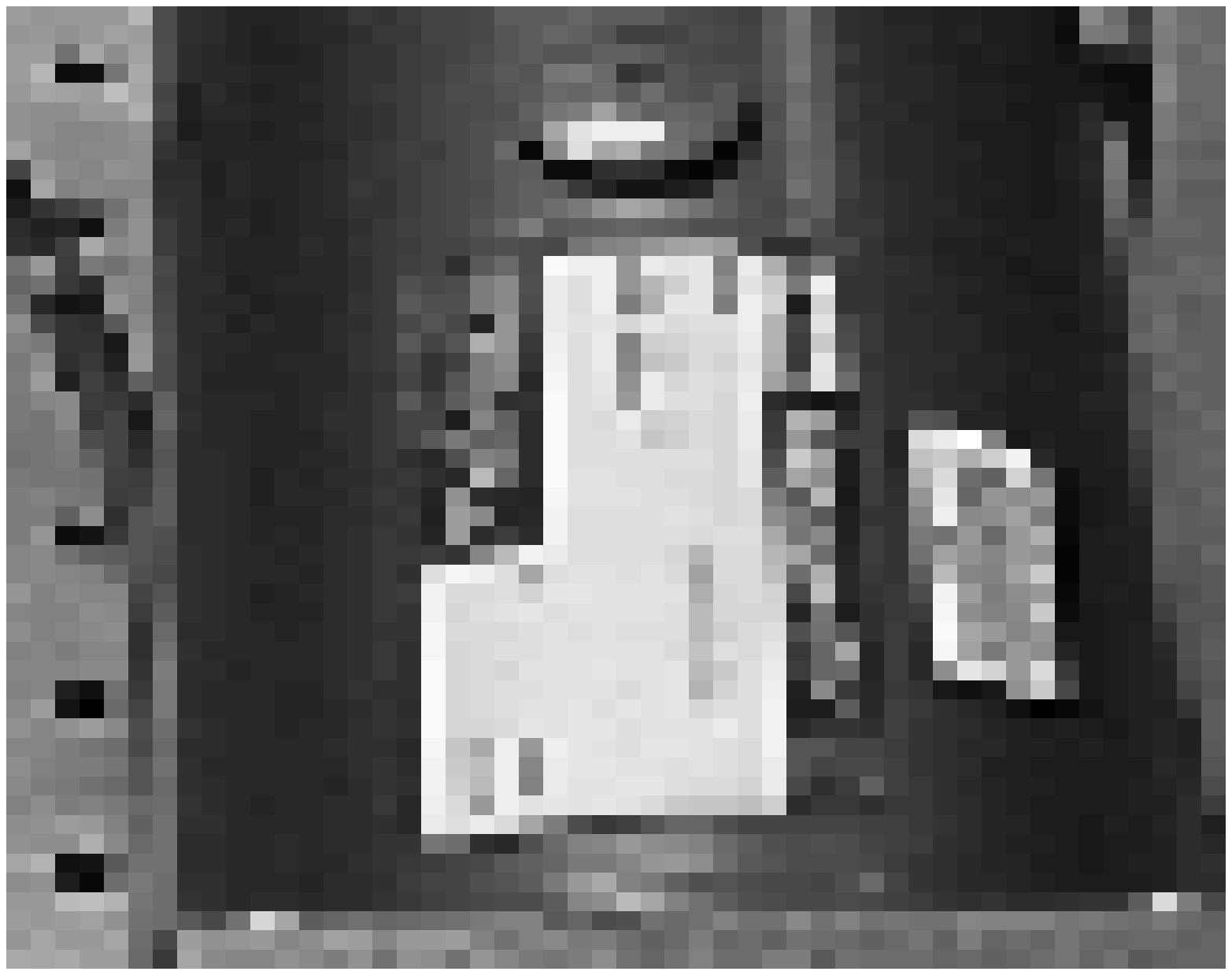,         width=1.60cm, height=1.60cm}
 \end{center}
 \caption
        {Test images: (a)~{\tt Build};
	 (b)~{\tt WoBA}; (c)~{\tt WoBB}; (d)~{\tt WoBC};
	 (e)~{\tt WoBD}; (f)~{\tt Cycl}; (g)~{\tt Sand};
	 (h)~{\tt ToolA}; (i)~{\tt ToolB}; (j)~{\tt ToolC}.
        }
 \label{fig:ImaDB}
\end{figure}

\begin{table}[ht]
  \small{
  \begin{center}
  \begin{tabular} {c||c|c||c|c}
    image & Int/0 &  Int/1.0 & Inv/0 & Inv/1.0 \\
    \hline 
     {\tt Build}   & 85.0 & 78.0 & 85.8 & 89.5 \\
     {\tt WoBA}    & 55.5 & 45.0 & 75.7 & 80.4 \\
     {\tt WoBB}    & 39.3 & 31.0 & 52.7 & 57.6 \\
     {\tt WoBC}    & 67.2 & 58.3 & 68.9 & 78.7 \\
     {\tt WoBD}    & 31.6 & 29.2 & 48.0 & 67.4 \\
     {\tt Cycl}    & 60.5 & 45.4 & 98.6 & 99.4 \\
     {\tt Sand}    & 50.5 & 40.9 & 85.2 & 94.4 \\
     {\tt ToolA}   & 41.7 & 35.3 & 60.2 & 68.0 \\
     {\tt ToolB}   & 29.5 & 23.4 & 45.7 & 54.1 \\
     {\tt ToolC}   & 42.1 & 27.8 & 42.5 & 48.4 \\
    \hline
     median        & 46.3 & 38.1 & 64.6 & 73.4 \\
     mean          & 50.3 & 41.4 & 66.3 & 73.8
  \end{tabular}
  \end{center}
  }
  \caption
        {Correlation accuracies CA, template size $6 \times 8$,
	 left columns for intensity data,
	 right columns for the invariant representation,
         without and with prefiltering.
        }
  \label{tab:CA}
\end{table}

\begin{figure}[hb]
 \begin{center}
  \epsfig{file=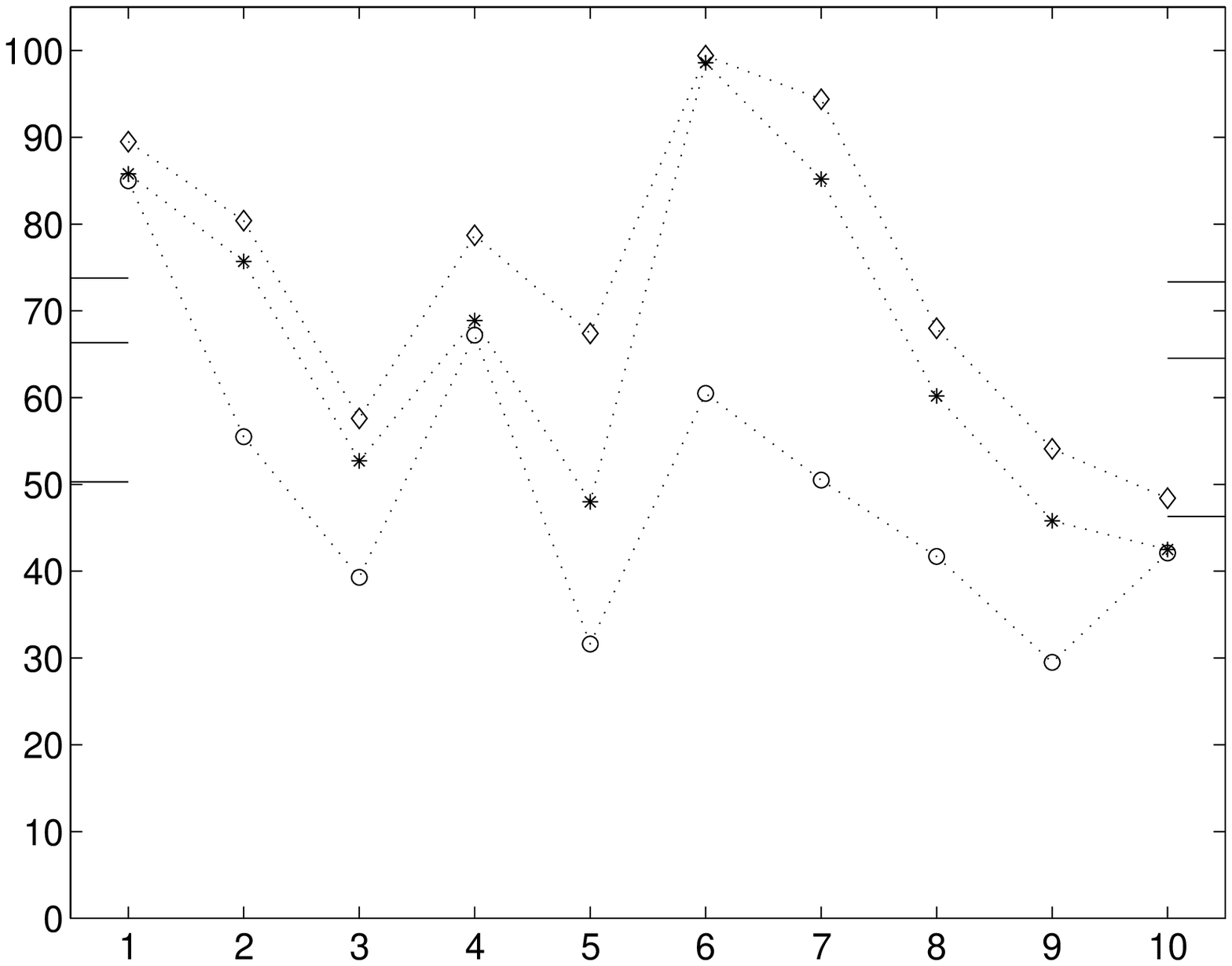,   width=8.0cm,height=4.5cm}
  \epsfig{file=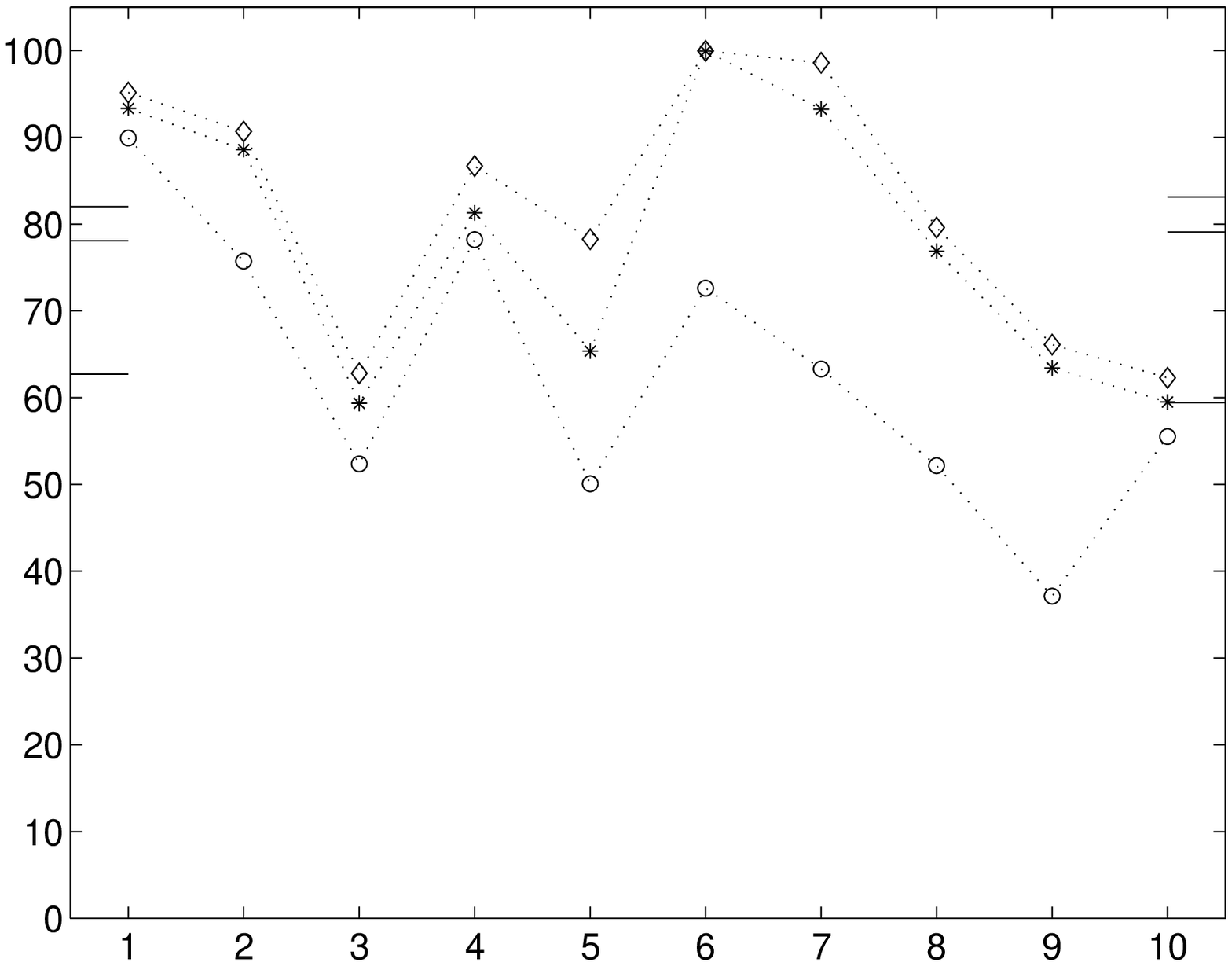, width=8.0cm,height=4.5cm}
  \caption
        {Correlation accuracies,
        \ (top) template size $6 \times 8$;
        \ (bottom) template size $10 \times 10$.
         The entries at x=1 refer to {\tt Build},
         at x=2 to {\tt WoBA}, etc.
        \ (circles, lower line) intensity images;
        \ (stars, center line) invariant representation, $\sigma_{pre}$=0;
        \ (diamonds, upper line) invariant representation, $\sigma_{pre}$=1.0.
        The markers on the left hand side indicate the means,
        the markers on the right hand side the medians.
        }
  \label{fig:CorrelAccuras}
 \end{center}
\end{figure}

\clearpage

We have computed the correlation accuracy for all the images given in
fig.~\ref{fig:ImaDB}. The results are shown in table~\ref{tab:CA}
and visualized in fig.~\ref{fig:CorrelAccuras}.
We observe the following:
\begin{itemize}
  \item The correlation accuracy CA is higher on the invariant representation
        than on the intensity images.
  \item The correlation accuracy  is higher on the invariant representation
        with gentle prefiltering, $\sigma_{pre}=1.0$, than without prefiltering.        We also observed a decrease in correlation accuracy if we increase
        the prefiltering well beyond $\sigma_{pre}=1.0$.
        By contrast, prefiltering seems to be always detrimental
        to the intensity images CA.
  \item The correlation accuracy shows a wide variation,
        roughly in the
        range 30\%$\dots$90\% for the unfiltered intensity images
        and 50\%$\dots$100\% for prefiltered invariant representations.
	Similarly, the gain in correlation accuracy
        ranges from close to zero up to 45\%.
        For our test images, it turns out that the invariant representation
        is always superior, but that doesn't necessarily have to be the case.
  \item The medians and means of the CAs over all test images
        confirm the gain in correlation accuracy
        for the invariant representation.
  \item The larger the template size, the higher the correlation accuracy,
        independent of the representation.
        A larger template size means more structure,
        and more discriminatory power.
\end{itemize}

%****************************************************************************
\section{Conclusion}
%****************************************************************************

We have proposed novel invariants that combine invariance
under gamma correction with invariance under geometric transformations.
In a general sense, the invariants can be seen as trading off derivatives
for a power law parameter, which makes them interesting for
applications beyond image processing.
The error analysis of our implementation on real images has shown
that, for sampled data, the invariants cannot be computed robustly everywhere.
Nevertheless, the template matching application scenario has demonstrated
that a performance gain is achievable by using the proposed invariant.

%****************************************************************************
\section{Acknowledgements}
%****************************************************************************

Bob Woodham suggested to the author to look into invariance
under gamma correction.
His meticulous comments on this work were much appreciated.
Jochen Lang helped with the acquisition of image data through
the ACME facility~\cite{pllw99}.

%****************************************************************************


\begin{thebibliography}{99}
%****************************************************************************

\bibitem{aw99}
R.~Alferez, Y.~Wang,
``Geometric and Illumination Invariants for Object Recognition'',
{\em IEEE Transactions on Pattern Analysis and Machine Intelligence},
Vol.21, No.6, pp.505-537, June 1999.

\bibitem{fmzchr91}
D.~Forsyth, J.~Mundy, A.~Zisserman, C.~Coelho, C.~Rothwell,
``Invariant Descriptors for 3-D Object Recognition and Pose'',
{\em IEEE Transactions on Pattern Analysis and Machine Intelligence},
Vol.13, No.10, pp.971-991, Oct.1991.

\bibitem{fua93}
P.~Fua,
``A parallel stereo algorithm that produces dense depth maps
  and preserves image features'',
{\em Machine Vision and Applications},
Vol.6, pp.35-49, 1993.

\bibitem{h98b}
G.~Holst,
``CCD arrays, cameras, and displays'',
JCD Publishing \& SPIE Press, 2nd ed., 1998.

\bibitem{horn86}
B.~Horn,
``Robot Vision'',
MIT Press, 1986.

\bibitem{mar93}
G.~Martin,
``High-Resolution Color CCD Camera'',
{\em Proc. SPIE Conference on Cameras, Scanners, and Image Acquisition Systems},pp.120-135, San Jose 1993.

\bibitem{mz92a}
J.~Mundy, A.~Zisserman,
``Geometric Invariance in Computer Vision'',
MIT Press, Cambridge 1992.

\bibitem{pllw99}
D.~Pai, J.~Lang, J.~Lloyd,~R. Woodham,
``ACME, A Telerobotic Active Measurement Facility'',
Sixth International Symposium on Experimental Robotics,
Sydney, 1999.
See also: http://www.cs.ubc.ca/nest/lci/acme/

\bibitem{poyn96}
C.~Poynton,
``A Technical Introduction to Digital Video'',
{\em John Wiley \& Sons}, 1996.
See also http://www.inforamp.net/~poynton/GammaFAQ.html.

\bibitem{rw95}
E.~Rivlin, I.~Weiss,
``Local Invariants for Recognition'',
{\em IEEE Transactions on Pattern Analysis and Machine Intelligence},
Vol.17, No.3, pp.226-238, Mar.1995.

\bibitem{sm97}
C.~Schmid, R.~Mohr,
``Local grayvalue invariants for Image Retrieval'',
{\em IEEE Transactions on Pattern Analysis and Machine Intelligence},
Vol.19, No.5, pp.530-535, May 1997.

\bibitem{s99}
A.~Siebert,
``A Differential Invariant for Zooming'',
{\em ICIP-99}, Kobe 1999.

\bibitem{r94}
B.~ter~Haar~Romeny,
``Geometry-Driven Diffusion in Computer Vision'',
Kluwer 1994.

\bibitem{w93}
I.~Weiss,
``Geometric Invariants and Object Recognition'',
{\em International Journal of Computer Vision},
Vol.10, No.3, pp.207-231, 1993.

\bibitem{wood96}
J.~Wood,
``Invariant Pattern Recognition: A Review'',
{\em Pattern Recognition},
Vol.29, No.1, pp.1-17, 1996.


\end{thebibliography}
\end{document}